\documentclass[runningheads]{llncs}
\usepackage[T1]{fontenc}
\usepackage{graphicx}
\newif\ifblindreview
\blindreviewfalse    

\usepackage{wrapfig} 
\usepackage{booktabs}
\usepackage{amsmath}
\usepackage{mathtools}
\usepackage{kotex}
\usepackage{graphicx}
\usepackage{multirow}
\usepackage[x11names]{xcolor}
\usepackage{bm}
\usepackage{hyperref}
\usepackage{url}

\usepackage{amsfonts}
\usepackage{mathtools}
\usepackage{subcaption}
\usepackage{algorithm}
\usepackage{algpseudocode}
\usepackage[most]{tcolorbox}
\usepackage{colortbl}
\usepackage{enumitem}

\makeatletter
\renewcommand\section{\@startsection{section}{1}{\z@}%
  {3.5ex \@plus 1ex \@minus 0.3ex}
  {1.5ex \@plus 0.5ex}
  {\normalfont\large\bfseries}}

\renewcommand\subsection{\@startsection{subsection}{2}{\z@}%
  {2.5ex \@plus 0.8ex \@minus 0.2ex}
  {1ex \@plus 0.3ex}
  {\normalfont\normalsize\bfseries}}

\renewcommand{\subsubsection}[1]{%
  \par\vspace{1ex}
  \noindent\textbf{#1}\;\ignorespaces
}
\makeatother

\newcommand{\bz}{{\bf z}}

\newcommand{\balpha}{\mbox{\boldmath{$\alpha$}}}

\newcommand{\bbeta}{\mbox{\boldmath{$\beta$}}}

\newcommand{\bc}{\begin{center}}
\newcommand{\ec}{\end{center}}
\newcommand{\be}{\begin{equation}}
\newcommand{\ee}{\end{equation}}
\newcommand{\ba}{\begin{array}}
\newcommand{\ea}{\end{array}}
\newcommand{\bean}{\begin{eqnarray*}}
\newcommand{\eean}{\end{eqnarray*}}
\newcommand{\bea}{\begin{eqnarray}}
\newcommand{\eea}{\end{eqnarray}}
\newcommand{\ben}{\begin{enumerate}}
\newcommand{\een}{\end{enumerate}}
\newcommand{\bed}{\begin{itemize}}
\newcommand{\eed}{\end{itemize}}

\newcommand{\beq}{\begin{equation}}
\newcommand{\eeq}{\end{equation}}
\newcommand{\beqn}{\begin{equation*}}
\newcommand{\eeqn}{\end{equation*}}

\newtheorem{assumption}{Assumption}

\begin{document}
\title{
Estimating Subgraph Importance with Structural Prior Domain Knowledge 
}

\titlerunning{eXEL: Subgraph Importance Estimation}
%

\ifblindreview
    \author{Anonymous Authors}
    \institute{Paper under double-blind review}
\else
    \author{
    Changhyun Kim\inst{1}\thanks{Both authors contributed equally to the paper.}
    \and
    Seunghwan An\inst{2}$^\star$
    \and
    Jong-June Jeon\inst{1}\thanks{Corresponding author.}
    }

    \authorrunning{Kim et al.}
    %
    \institute{
    University of Seoul, Seoul, South Korea \\
    \email{\{ch.kim, jj.jeon\}@uos.ac.kr}
    \and
    Incheon National University, Incheon, South Korea \\
    \email{sh.an@inu.ac.kr}}
\fi

\maketitle              
\begin{abstract}
We propose a subgraph importance estimation method for pretrained Graph Neural Networks (GNNs) on graph-level tasks, formulated as a linear Group Lasso regression problem in the embedding space. Our method effectively leverages prior domain knowledge of graph substructures, while remaining independent of the specific form of the output layer or readout function used in the GNN architecture, and it does not require access to ground-truth target labels. Experiments on real-world graph datasets demonstrate that our method consistently outperforms existing baselines in subgraph importance estimation. Furthermore, we extend our method to identify important nodes within the graph.
\keywords{Subgraph importance estimation \and Graph Neural Networks \and Linear Group Lasso regression \and Embedding space \and Graph mining.}
\end{abstract}


\section{Introduction} \label{sec:intro}

Graph Neural Networks (GNNs) have emerged as a specialized model for analyzing graph-structured data due to their ability to effectively capture both node features and the connectivity relationships between nodes via edges \cite{HGexplainer,graph_lime,graph_class_survey}. 
Specifically, the post-hoc approaches providing an interpretation of graphs via fitted node feature matrices have attracted considerable attention due to their simplicity and independence from specific model architectures.

In this context, GraphLIME \cite{graph_lime} and STGraphLIME \cite{stgraphlime} propose local explanation methods that identify influential nodes or subgraphs based on the Hilbert-Schmidt Independence Criterion (HSIC) Lasso \cite{hsic}. SubgraphX \cite{subgraphx} extends the Shapley value \cite{shapley} to the graph domain by computing the contributions of connected subgraphs, thereby providing subgraph-level explanations for GNN predictions. The aforementioned models retrospectively estimate node importance by quantifying changes in the GNN's outputs with respect to variations in graph-level features. 

However, existing explanation methods for GNN models are limited in their ability to incorporate prior knowledge, resulting in the frequent identification of important subgraphs that are not accepted by domain experts. For instance, subgraphs such as ecological clusters, functional components in engineered systems, and functional groups or motifs in molecular structures are known to determine key graph-level properties \cite{chemical_motif,subgraphx}. This observation indicates that subgraph-level explanations can provide a more domain-aligned understanding of model behavior. Furthermore, the permutation-invariant readout function in GNNs often makes the contribution of each node embedding to the final network output identical, which hinders the development of gradient-based methods.

To address these two issues of post-hoc GNN explanation methods, we propose a simple and effective domain-aligned important subgraph identification method based on a structural sparsity prior, \textit{Group Lasso} \cite{group_lasso}. Our method can incorporate domain-specific structural knowledge through the predefined groups. In addition, since it is independent of the GNN architecture (such as the readout function and output layer), it can be easily applied to various GNN models, aiming to evaluate the importance of substructures of a graph. To evaluate our method's effectiveness, we employ 13 real-world graph datasets focusing on molecular property prediction, comparing the results against ground-truth where available. Our method is referred to as \underline{eX}plainable \underline{E}mbedding-Space Group \underline{L}asso regression for GNN ({\textbf{eXEL}}\footnote{Our implementation code and Appendix are available at:\\\url{https://github.com/Nempet9398/graph_xai}.}).

\section{Related Works}


Existing approaches for GNN explanation and importance estimation can be categorized into four groups: surrogate-, decomposition-, perturbation-, and gradient-based methods.
Surrogate methods fit simple proxy models to the local decision boundary of a GNN; for example, GraphLIME \cite{graph_lime} applies kernel regression with HSIC Lasso on a node's neighborhood to identify important features, RelEx \cite{relex} trains a Graph Convolution Network (GCN) surrogate on sampled subgraphs to learn explanation masks, and PGM-Explainer \cite{pgm_explainer} captures dependencies between features and predictions using a probabilistic graphical model. 
Decomposition methods redistribute the model's output to input components. LRP \cite{explain_technique} propagates relevance scores through layers, ExcitationBP \cite{explain_method} redistributes relevance via conditional likelihood, and GNN-LRP \cite{GNN_LRP} attributes importance along multi-hop walks via higher-order Taylor expansion.
Perturbation-based and gradient-based methods estimate importance by modifying inputs or analyzing gradients. 
Perturbation methods such as GNNExplainer \cite{gnn_explainer}, PGExplainer \cite{pg_expaliner}, GraphMask \cite{ig_gnn}, and SubgraphX \cite{subgraphx} assess the sensitivity of predictions to feature or edge changes.
Gradient- and feature-based methods, including Saliency Analysis, Guided Backpropagation, CAM, and Grad-CAM \cite{explain_technique,explain_method}, estimate importance directly from gradients or feature maps. 

\section{Preliminaries}

\subsubsection{Notations.}
Consider a graph consisting of $n$ nodes, denoted by $G = (X, A)$, where $X$ is the node feature matrix whose rows represent node attributes, and $A$ is the adjacency matrix. A labeled dataset is given by $\mathcal{D} = \{(G^{(i)}, y^{(i)})\}_{i=1}^N = \{(X^{(i)}, A^{(i)}, y^{(i)})\}_{i=1}^N$. To represent a subgraph of $G$, we introduce $\mathcal{P}_G$, an $m$-partition of the node index set $\{1, 2, \ldots, n\}$, and let $I_s \in \mathcal{P}_G$ ($s = 1, 2, \ldots, m$) denote the index set of nodes belonging to the $s$th subgraph. 


\subsubsection{Graph Neural Networks.}
Let the composition of the embedding layer and the graph convolution function be denoted by $f$, which outputs the embedding matrix of a given graph $G$ lying in $\mathbb{R}^d$ as follows: 
$f: \mathbb{R}^{n \times p} \times \{0, 1\}^{n \times n} \mapsto \mathbb{R}^{n \times d}$.
Here, the first dimension, corresponding to the number of nodes, remains invariant. 
The readout function, denoted as $h$, produces a $d$-dimensional graph-level representation by applying a row-wise permutation-invariant aggregation to the output of $f$, such as a column-wise mean, sum, or maximum (we will demonstrate in Section \ref{sec:exp} that our proposed method is applicable to all these cases). Lastly, the final output layer $h_{\text{out}}: \mathbb{R}^{d} \mapsto \mathcal{Y}$ determines the form of the output space $\mathcal{Y}$ (e.g., class probabilities for classification tasks).
Then, the GNN model is trained by the following objective:
\begin{footnotesize}
\beq \label{eq:optim}
\min_{h_{\text{out}}, f} \quad \frac{1}{N} \sum_{i=1}^N \mathcal{L}\Big( h_{\text{out}} \big( (h \circ f)(X^{(i)}, A^{(i)}) \big), \, y^{(i)} \Big),
\eeq
\end{footnotesize}
where $\mathcal{L}: \mathcal{Y} \times \mathcal{Y} \mapsto \mathbb{R}^+$ is the loss function corresponding to the target task (e.g., cross-entropy loss for classification), and $y^{(i)} \in \mathcal{Y}$ denotes the target label.

\section{Proposal: eXEL} \label{sec:proposal}

Our goal is to develop an importance estimation method that assigns continuous importance scores to all subgraphs for a GNN model, which is pretrained according to \eqref{eq:optim}. 
We introduce the concept of subgraph-level sparsity and define importance as the norm of the subgraph feature.

\subsection{Motivation} \label{sec:rationale}

\begin{wrapfigure}{r}{0.55\linewidth}
\vspace{-14mm}
\begin{subfigure}[b]{0.48\linewidth}
    \includegraphics[width=\linewidth]{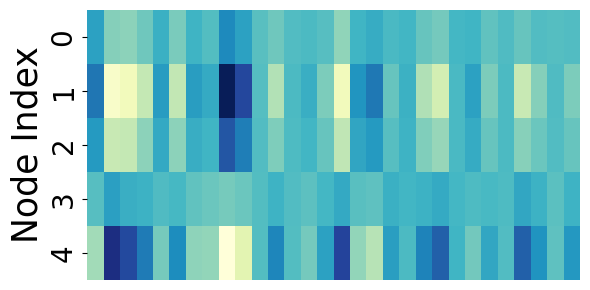}
    \caption{Node embeddings}
\end{subfigure}
\hfill
\begin{subfigure}[b]{0.48\linewidth}
    \includegraphics[width=\linewidth]{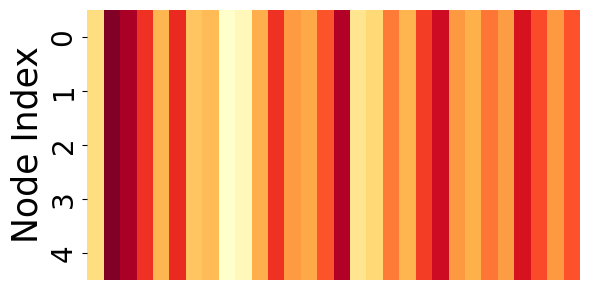}
    \caption{Gradients}
\end{subfigure}
\caption{
(a): Node embedding matrix.
(b): Gradient values with respect to the node embeddings.
}
\label{fig:gradient_pooling}
\vspace{-8mm}
\end{wrapfigure}

In GNN, the permutation-invariant readout function is widely used to aggregate node features into a fixed-size graph-level representation. While this readout function preserves the isomorphic property of graphs, the node contributions are pooled, which hinders the explanation of graphs on the node level. Figure \ref{fig:gradient_pooling} (a) and (b) show the node embedding matrix and the gradient of the node embeddings. Since all gradient vectors for each node embedding are equal, it is difficult to directly apply gradient-based attribution methods.

Regardless of the pretraining task (i.e., independent of the choice of the output layer $h_{\text{out}}$ or the output space $\mathcal{Y}$), the graph-level embedding, $(h \circ f)(X, A)$, serves as the \textit{optimal} $d$-dimensional representation of the given graph, in the sense that the pretrained GNN model achieves its best performance on the graph-level task using this embedding vector. 
In addition, since the readout function aggregates node embeddings along each embedding dimension, a mapping exists for the same graph-level representations when the number of nodes is large.
If the prior knowledge restricts a partition of subgraphs, $\mathcal{P}_G$, the same graph-level representation within the constraint can be reasonably interpreted in the context of a specific data domain.

Thus, our approach is guided by the motivation that the explanation for a graph should be as structurally simple as possible under the prior domain knowledge $\mathcal{P}_G$. Let $\balpha = (\alpha_1, \cdots, \alpha_n)^\top$ and a subvector of $\balpha$ be $\balpha_s = (\alpha_k: k \in I_s \in \mathcal{P}_G)$, then our ideal objective function is given by $\max_{\alpha \in \mathbb{R}^n} \sum_{s = 1}^m \mathbb{I} \big(\|\balpha_s\| = 0\big)$ with the constraint $(h \circ f)(X, A) = \balpha^\top f(X, A)$. 
Since this constraint is generally infeasible for non-linear readout functions, we set our objective function by 
\begin{footnotesize}
\bea \label{eq:ideal}
\max_{\balpha\in \mathbb{R}^n} \quad \sum_{s = 1}^m \mathbb{I} \big(\|\balpha_s\| = 0\big) \quad
\mbox{subject to } \big\|(h \circ f)(X, A) - \balpha^\top f(X, A)\big\|<\epsilon,
\eea
\end{footnotesize}
where $\mathbb{I}(\cdot)$ is indicator function, $\|\cdot\|$ denotes the $L_2$ norm, and $\epsilon$ is a positive tuning parameter.
The constraint in \eqref{eq:ideal} can be interpreted as a well-posed regression problem when $d > n$, where $d$ and $n$ correspond to the number of samples and the number of covariates in linear regression framework, respectively (in our experiments, we set $d = 256$, and the average number of nodes was $26$; see the Appendix for detailed dataset descriptions).
In the evaluation, we used mean and sum readout functions, which are commonly employed for molecular graph datasets \cite{adaptive_readout}, as well as the max operation for a dimension-wise nonlinear readout function.

\begin{figure}[t]
    \centering
    \includegraphics[width=0.9\linewidth]{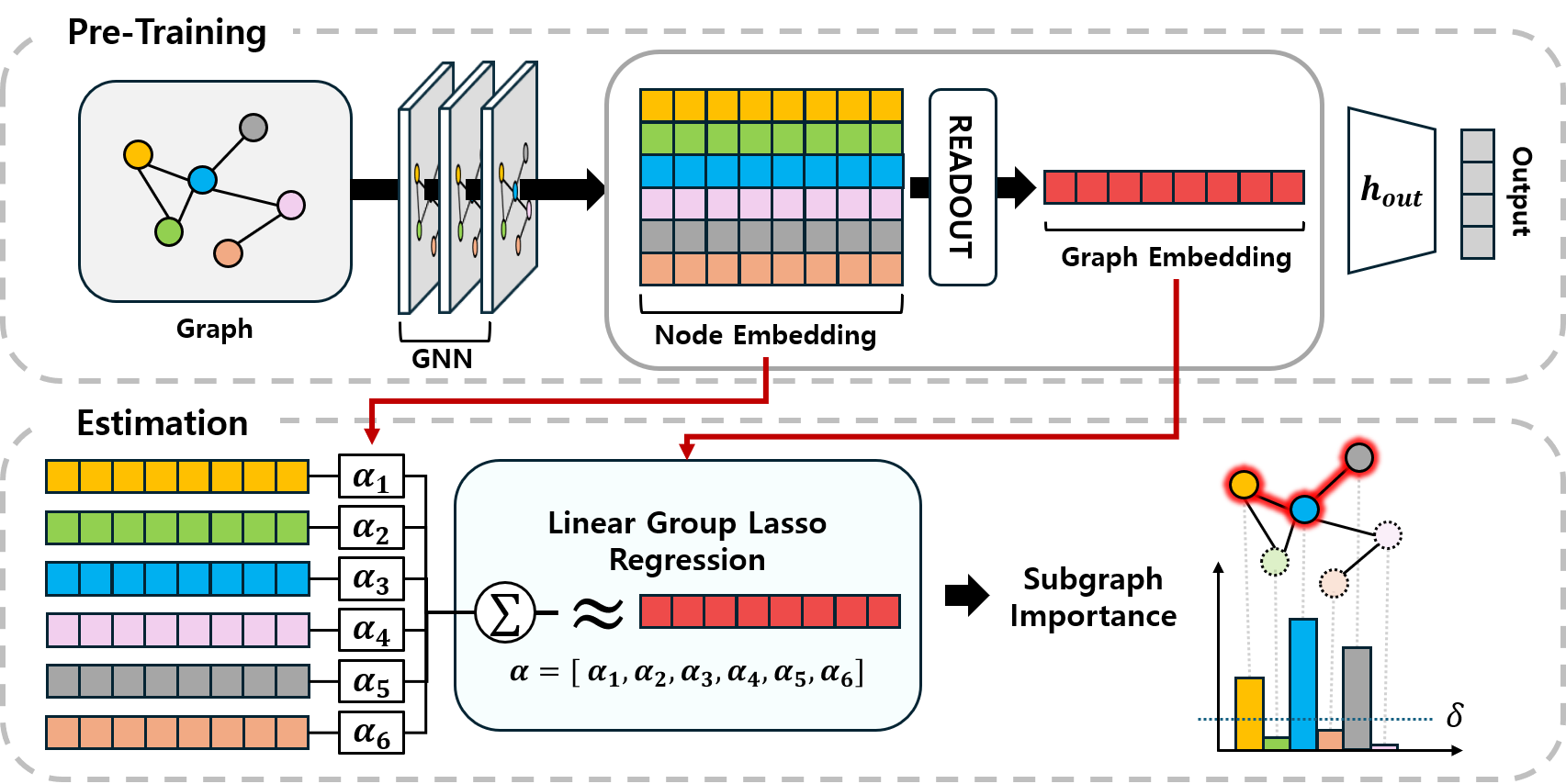}
    \caption{Overall process of our proposed subgraph importance estimation method.}
    \label{fig:overall}
\end{figure} 


\subsection{Subgraph Importance Estimation}

Since \eqref{eq:ideal} is a nonconvex combinatorial problem, we relax this objective function using the Group Lasso penalty and define, for each graph, the corresponding optimization problem as follows:
\begin{footnotesize}
\beq \label{eq:grouplasso}
\min_{\balpha \in \mathbb{R}^n} \quad \big\|  (h \circ f)(X, A) - \balpha^\top f(X, A)  \big\|^2 + \lambda \sum_{s=1}^m \| \balpha_{s} \|,
\eeq
\end{footnotesize}
where $\lambda > 0$ is a hyperparameter that controls the degree of sparsity.  
From \eqref{eq:grouplasso}, we can observe the following advantages:
\ben
    \item Domain knowledge of graph substructures is incorporated into the sparsity penalty by grouping nodes that belong to each known subgraph unit.
    \item Our method is independent of the GNN architecture, including the readout function, output layer, and target space.
\een

\begin{assumption} \label{assump:lipschitz}
The loss function $\mathcal{L}$ used for the pre-training of a GNN model is $K$-Lipschitz with respect to its first argument if, for all graph-level embedding vectors $\bz_1, \bz_2 \in \mathbb{R}^d$ and $y \in \mathcal{Y}$, $\big\vert \mathcal{L}( \bz_1, y ) - \mathcal{L}( \bz_2, y ) \big\vert \leq K \| \bz_1 - \bz_2 \|$, where $K > 0$.
\end{assumption}

Without loss of generality, the loss function $\mathcal{L}$ in \eqref{eq:optim} can be rewritten for a fixed GNN model as $\mathcal{L}( (h \circ f)(X, A), y )$, where the output layer $h_{\text{out}}$ is considered to be absorbed into the loss function $\mathcal{L}$. 
Note that many loss functions commonly used in graph-level tasks (e.g., cross-entropy \cite{lip_bio,lip_bio_1,lip_network,lip_text_1}, hinge loss \cite{lip_hinge}) satisfy Assumption \ref{assump:lipschitz}.
Then, under Assumption \ref{assump:lipschitz},
\begin{footnotesize}
\beq \label{eq:lipschitz}
\Big\vert \mathcal{L}\big( (h \circ f)(X, A), y \big) - \mathcal{L}\big( \balpha^\top f(X, A), y \big) \Big\vert^2 \leq K^2 \Big\| (h \circ f)(X, A) - \balpha^\top f(X, A) \Big\|^2.
\eeq
\end{footnotesize}

Since our optimization problem \eqref{eq:grouplasso} is a well-posed linear regression problem with a large number of nodes, it can yield group coefficients $\balpha_s$ that produce nearly identical graph-level representations. 
Moreover, reliable $\balpha_s$ can be obtained under Assumption \ref{assump:lipschitz}, as \eqref{eq:lipschitz} ensures similar loss values (note that the right-hand side of \eqref{eq:lipschitz} corresponds to the first term in our objective \eqref{eq:grouplasso}). 
Therefore, our proposed method can structurally select subgraphs that yield the same model prediction, guided by prior domain knowledge.

After solving the optimization problem in \eqref{eq:grouplasso}, we transform the resulting score vector $\balpha$ to obtain the estimate of subgraph importance $\bbeta = (\beta_1,\cdots,\beta_m)$:
\begin{footnotesize}
\beqn
\beta_s \coloneqq \frac{1}{\vert I_s \vert} \sum_{l \in I_s} \alpha_l, \quad \text{for} \,\, s=1,2,\cdots,m,
\eeqn
\end{footnotesize}
which serves as an averaging step to ensure consistent importance values among nodes within the same subgraph.
Then, we define the set of important subgraphs $\mathcal{S}$ in graph $G$ as those whose absolute scores exceed a threshold $\delta > 0$, which filters out subgraphs with negligible weights (in practice, we set $\delta = 10^{-5}$):
\begin{footnotesize}
\beqn
\mathcal{S} \coloneqq \big\{ s \in \{1,2,\cdots,m\} \mid \delta < \vert \beta_s \vert \big\}.
\eeqn
\end{footnotesize}
The overall process of our proposed method is illustrated in Figure \ref{fig:overall}.

\section{Experiments} \label{sec:exp}

We evaluate eXEL by addressing the following research questions:
\ben
    \item \textbf{Estimation performance:} Can eXEL accurately identify important subgraphs?
    \item \textbf{Ablation study:} Does the Group Lasso penalty effectively incorporate domain-specific prior knowledge?
    \item \textbf{Model analysis:} How does reconstructing the original graph-level embedding contribute to importance estimation?
    \item \textbf{Scalability:} Can eXEL be extended to node-level importance estimation?
\een

\subsection{Overview}

\subsubsection{Datasets.}
Following prior studies \cite{gnn_explainer,subgraphx}, we evaluate our method on molecular datasets that exhibit interpretable subgraph structures such as functional groups. We use \texttt{MUTAG} \cite{tud}, \texttt{Benzene}(\texttt{BENZ}), \texttt{Alkane}, \texttt{Fluoride}(\texttt{FLUOR}) \cite{graphxai,gt_dataset}, \texttt{Ames} \cite{tdc}, \texttt{BACE}, and \texttt{BBBP} \cite{ogb}, where atomic features represent node attributes. Motif structures (i.e., predefined groups) are extracted using BRICS decomposition \cite{chemical_motif} and tree decomposition \cite{motif_ssl}. Among these, \texttt{Benzene}, \texttt{Alkane}, and \texttt{Fluoride} provide ground-truth annotations for subgraph-level importance. Each dataset is randomly split into 80\% training and 20\% test, and the optimal $\lambda$ is selected via 4-fold cross-validation.

\subsubsection{Baselines.}
Following prior work \cite{subgraphx}, we train four widely used GNN architectures: GCN, GAT, GIN, and GraphSAGE (SAGE). We also experiment with three types of readout functions (mean, sum, and max).
Each GNN architecture is trained using the cross-entropy loss, which is Lipschitz continuous (i.e., Assumption \ref{assump:lipschitz}), and the following baseline explanation methods are applied to the fitted GNNs: 
PGExplainer (PGExp) \cite{pg_expaliner}, GNNExplainer (GNNExp) \cite{gnn_explainer}, GradCAM \cite{explain_method}, GraphMask \cite{ig_gnn}, Guided Backpropagation (GBP) \cite{explain_method}, Saliency Maps (SA) \cite{explain_method}, and SubgraphX \cite{subgraphx}.
Each method generates importance scores over nodes and averages them to provide a subgraph-level explanation.
Further details about the baseline models and configurations are provided in the Appendix.

\subsubsection{Evaluation.}
We adopt a \textit{fidelity metric} to measure the influence of the identified subgraphs on model predictions. Extending the metric proposed by \cite{fidelity}, we compute the drop in $\mathrm{F_1}$-score as $\texttt{Fidelity}_{\mathrm{F_1}} = \mathrm{F_1}(G) - \mathrm{F_1}(G \setminus S)$, where $S$ is the selected subset of nodes, $\mathrm{F_1}(G)$ denotes the $\mathrm{F_1}$-score on the original graph, and $\mathrm{F_1}(G \setminus S)$ is obtained by zeroing out the features of nodes in $S$. 
A higher fidelity score indicates that the identified subset of nodes has a stronger influence, i.e., greater importance.
All fidelity scores are averaged over different readout functions.

\begin{table}[t]
\scriptsize
\centering
\caption{
Subgraph-level $\texttt{Fidelity}_{\mathrm{F_1}}$ scores. The best result is highlighted with \colorbox{gray!20}{color} and the second-best is \underline{underlined}. Higher is better.
}
\setlength{\tabcolsep}{0.5mm}
\resizebox{0.99\textwidth}{!}{
\begin{tabular}{ll|c|cccccccc}
\toprule
model & data & eXEL & PGExp & GNNExp & GradCAM & GraphMask & GBP & SA & SubgraphX \\
\midrule
\multirow{7}{*}{GCN}
 & \texttt{MUTAG} & 
 $\underline{.158_{\pm .140}}$ & $.000_{\pm{.094}}$ & $.151_{\pm{.122}}$ & $.001_{\pm{.068}}$ & $.009_{\pm{.075}}$ & $.031_{\pm{.181}}$ & $.020_{\pm{.121}}$ & $\cellcolor{gray!20}{.243_{\pm{.297}}}$ \\
 & \texttt{alkane} & $\cellcolor{gray!20}{.325_{\pm{.138}}}$ & $.258_{\pm{.192}}$ & $.277_{\pm{.180}}$ & \underline{$.321_{\pm{.191}}$} & $.097_{\pm{.094}}$ & $.133_{\pm{.127}}$ & $.225_{\pm{.111}}$ & $.080_{\pm{.086}}$ \\
 & \texttt{ames} & $\underline{.155_{\pm{.155}}}$ & $.024_{\pm{.044}}$ & $.082_{\pm{.057}}$ & $.009_{\pm{.039}}$ & $.037_{\pm{.042}}$ & $.026_{\pm{.032}}$ & $.044_{\pm{.070}}$ & $\cellcolor{gray!20}{.194_{\pm{.298}}}$ \\
 & \texttt{bace} & $\cellcolor{gray!20}{.243_{\pm .038}}$ &
 $.088_{\pm{.051}}$ & \underline{$.165_{\pm{.057}}$} & {$.136_{\pm{.058}}$} & $.071_{\pm{.049}}$ & $.116_{\pm{.053}}$ & $.119_{\pm{.070}}$ & $.108_{\pm{.062}}$ \\
 & \texttt{bbbp} & $\cellcolor{gray!20}{.090_{\pm{.041}}}$ & $.041_{\pm{.030}}$ & \underline{$.069_{\pm{.026}}$} & $.044_{\pm{.054}}$ & $.043_{\pm{.034}}$ & $.054_{\pm{.034}}$ & $.068_{\pm{.028}}$ & $.055_{\pm{.045}}$ \\
 & \texttt{BENZ} & $\cellcolor{gray!20}{.196_{\pm{.112}}}$ & $.076_{\pm{.130}}$ & \underline{$.156_{\pm{.044}}$} & $.075_{\pm{.060}}$ & $.073_{\pm{.031}}$ & $.147_{\pm{.080}}$ & $.095_{\pm{.056}}$ & $.112_{\pm{.034}}$ \\
 & \texttt{FLUOR} & $\cellcolor{gray!20}{.323_{\pm{.281}}}$ & $.171_{\pm{.303}}$ & $.142_{\pm{.183}}$ & $.122_{\pm{.084}}$ & $.105_{\pm{.120}}$ & $.217_{\pm{.185}}$ & $\underline{.218_{\pm{.316}}}$ & $.104_{\pm{.101}}$ \\

\midrule
\multirow{7}{*}{GAT}
& \texttt{MUTAG} & $\underline{.138_{\pm .152}}$ &
$.006_{\pm.059}$ & $.044_{\pm.063}$ & $.016_{\pm.052}$ & $.034_{\pm.042}$ & ${.058_{\pm.058}}$ & $.035_{\pm.198}$ & $\cellcolor{gray!20}{.142_{\pm.100}}$ \\
& \texttt{alkane} & $\underline{.455_{\pm .198}}$ &
$.164_{\pm.169}$ & $.336_{\pm.149}$ & $.409_{\pm.149}$ & $.144_{\pm.099}$ & $\cellcolor{gray!20}{.482_{\pm.269}}$ & $.073_{\pm.074}$ & $.073_{\pm.306}$ \\
& \texttt{ames} & $\cellcolor{gray!20}{.168_{\pm .046}}$ &
$.074_{\pm.053}$ & \underline{$.146_{\pm.014}$} & $.035_{\pm.019}$ & $.072_{\pm.041}$ & $.100_{\pm.053}$ & $.099_{\pm.046}$ & $.014_{\pm.039}$ \\
& \texttt{bace} & $\cellcolor{gray!20}{.247_{\pm .055}}$ &
$.121_{\pm.107}$ & \underline{$.222_{\pm.075}$} & $.123_{\pm.080}$ & $.158_{\pm.068}$ & $.188_{\pm.107}$ & $.149_{\pm.062}$ & $.131_{\pm.073}$ \\
& \texttt{bbbp} & $\cellcolor{gray!20}{.079_{\pm.036}}$ & $.036_{\pm.025}$ & \underline{$.068_{\pm.027}$} & $.026_{\pm.017}$ & $.049_{\pm.037}$ & $.043_{\pm.025}$ & $.043_{\pm.027}$ & $.043_{\pm.027}$ \\
& \texttt{BENZ} & $\cellcolor{gray!20}{.578_{\pm.144}}$ & $.062_{\pm.050}$ & $.163_{\pm.049}$ & $.045_{\pm.054}$ & $.105_{\pm.076}$ & $.166_{\pm.049}$ & $.158_{\pm.058}$ & $\underline{.558_{\pm.144}}$ \\
& \texttt{FLUOR} & $\cellcolor{gray!20}{.587_{\pm.182}}$ & $.177_{\pm.232}$ & $.316_{\pm.135}$ & $.317_{\pm.232}$ & $.356_{\pm.099}$ & $.517_{\pm.233}$ & $.244_{\pm.216}$ & $\underline{.538_{\pm.182}}$ \\

\midrule
\multirow{7}{*}{GIN}
& \texttt{MUTAG} & $\underline{.139_{\pm .175}}$ & 
$.005_{\pm.027}$ & $.010_{\pm.162}$ & $.095_{\pm.210}$ & $.006_{\pm.090}$ & $.013_{\pm.047}$ & $.074_{\pm.138}$ & $\cellcolor{gray!20}{.141_{\pm.251}}$ \\
& \texttt{alkane} & $\underline{.406_{\pm.370}}$ & $.158_{\pm.325}$ & $.147_{\pm.174}$ & $\cellcolor{gray!20}{.424_{\pm.372}}$ & $.112_{\pm.078}$ & $.246_{\pm.241}$ & $.381_{\pm.245}$ & $.041_{\pm.065}$ \\
& \texttt{ames} & $\cellcolor{gray!20}{.242_{\pm .060}}$ & 
$.062_{\pm.045}$ & $.109_{\pm.050}$ & $.144_{\pm.086}$ & $.103_{\pm.041}$ & $.100_{\pm.054}$ & \underline{$.176_{\pm.089}$} & $.174_{\pm.112}$ \\
& \texttt{bace} & $\cellcolor{gray!20}{.228_{\pm.081}}$ & 
$.089_{\pm.064}$ & $.206_{\pm.066}$ & $.218_{\pm .064}$ & $.112_{\pm.060}$ & $.185_{\pm.063}$ & $\underline{.216_{\pm.109}}$ & $.162_{\pm.073}$ \\
& \texttt{bbbp} & $\cellcolor{gray!20}{.111_{\pm.047}}$ & $.032_{\pm.023}$ & \underline{$.061_{\pm.016}$} & $.032_{\pm.021}$ & $.047_{\pm.017}$ & $.032_{\pm.021}$ & $.036_{\pm.026}$ & $.034_{\pm.035}$ \\
& \texttt{BENZ} & $\cellcolor{gray!20}{.388_{\pm .095}}$ & 
$.057_{\pm.066}$ & $.160_{\pm.087}$ & $.206_{\pm.162}$ & $.075_{\pm.016}$ & $.152_{\pm.088}$ & \underline{$.331_{\pm.088}$} & $.168_{\pm.053}$ \\
& \texttt{FLUOR} & $\underline{.596_{\pm .190}}$ &
$.227_{\pm.284}$ & $.399_{\pm.194}$ & $\cellcolor{gray!20}{.598_{\pm.212}}$ & $.216_{\pm.115}$ & $.439_{\pm.275}$ & $\cellcolor{gray!20}{.598_{\pm.321}}$ & $.146_{\pm.180}$ \\

\midrule
\multirow{7}{*}{SAGE}
& \texttt{MUTAG} & $\underline{.200_{\pm .253}}$ & 
$.012_{\pm.039}$ & $.075_{\pm.091}$ & $.091_{\pm.205}$ & $.031_{\pm.041}$ & $.038_{\pm.067}$ & $.060_{\pm.089}$ & $\cellcolor{gray!20}{.258_{\pm.312}}$ \\
& \texttt{alkane} & $\cellcolor{gray!20}{.587_{\pm .277}}$ & 
$.127_{\pm.204}$ & $.244_{\pm.125}$ & $\underline{.532_{\pm.357}}$ & $.112_{\pm.057}$ & $.294_{\pm.245}$ & $.352_{\pm.195}$ & $.078_{\pm.065}$ \\
& \texttt{ames} & $\cellcolor{gray!20}{.265_{\pm .089}}$ &
$.052_{\pm.027}$ & $.160_{\pm.073}$ & $.139_{\pm.105}$ & $.098_{\pm.098}$ &  $.147_{\pm.091}$ &
$\underline{.212_{\pm.068}}$ &
\underline{$.212_{\pm.069}$} \\
& \texttt{bace} & $\cellcolor{gray!20}{.240_{\pm.083}}$ & $.100_{\pm.047}$ & {$.198_{\pm.083}$} & $.192_{\pm.065}$ & $.132_{\pm.080}$ & \underline{$.217_{\pm.090}$} & $.160_{\pm.049}$ & $.160_{\pm.130}$ \\
& \texttt{bbbp} & $\cellcolor{gray!20}{.115_{\pm .076}}$ & 
$.044_{\pm.018}$ & $.060_{\pm.032}$ & $.027_{\pm.022}$ & $.037_{\pm.039}$ & \underline{${.061_{\pm.050}}$} & $.026_{\pm.046}$ & $.047_{\pm.017}$ \\
& \texttt{BENZ} & $\cellcolor{gray!20}{.370_{\pm.143}}$ & $.053_{\pm.053}$ & $.161_{\pm.085}$ & $.150_{\pm.150}$ & $.065_{\pm.024}$ & $.161_{\pm.127}$ & $\underline{.281_{\pm.159}}$ & $.149_{\pm.052}$ \\
& \texttt{FLUOR} & $\cellcolor{gray!20}{.665_{\pm .218}}$ & 
$.239_{\pm.265}$ & $.433_{\pm.162}$ & $.459_{\pm.311}$ & $.227_{\pm.085}$ & ${.480_{\pm.224}}$ & $\underline{.571_{\pm.266}}$ & $.206_{\pm.130}$ \\

\bottomrule
\end{tabular}
}
\label{tab:sub_exp}
\end{table}

\begin{table}[t]
\scriptsize
\centering
\caption{
\textbf{Ablation study}: Effect of the Group Lasso penalty on subgraph-level $\texttt{Fidelity}_{\mathrm{F_1}}$ scores.
`Improve(\%)' denotes the relative performance difference between the two penalties, and the {\colorbox{gray!20}{color}} entries highlight positive improvements.
}
\setlength{\tabcolsep}{0.3mm}
\resizebox{0.95\textwidth}{!}{
\begin{tabular}{ll|ccccccccc}
\toprule
model & penalty & \texttt{MUTAG} & \texttt{alkane} & \texttt{ames} & \texttt{bace} & \texttt{bbbp} & \texttt{BENZ} & \texttt{FLUOR} \\
\midrule
\multirow{3}{*}{GCN}
& Lasso & $.115_{\pm{.133}}$ & ${.473_{\pm .163}}$ & $.148_{\pm .057}$ & ${.209_{\pm{.046}}}$ & ${.091_{\pm .034}}$ & ${.180_{\pm .140}}$ & ${.217_{\pm .216}}$ \\
& GroupLasso & ${.158_{\pm .140}}$ & ${.325_{\pm{.138}}}$ & ${.155_{\pm{.155}}}$ & ${.243_{\pm .038}}$ & ${.090_{\pm{.041}}}$ & ${.196_{\pm{.112}}}$ & ${.323_{\pm{.281}}}$ \\
& Improve(\%) & {\cellcolor{gray!20}{$\mathbf{+37.4}$}} & $-31.3$ & {\cellcolor{gray!20}{$\mathbf{+4.7}$}} & {\cellcolor{gray!20}{$\mathbf{+16.3}$}} & $-1.1$ & {\cellcolor{gray!20}{$\mathbf{+8.9}$}} & {\cellcolor{gray!20}{$\mathbf{+48.8}$}} \\
\midrule
\multirow{3}{*}{GAT}
& Lasso & $.106_{\pm.100}$ & $.411_{\pm.106}$ & ${.152_{\pm.046}}$ & ${.225_{\pm.073}}$ & ${.088_{\pm .024}}$ & ${.538_{\pm .169}}$ & ${.484_{\pm .158}}$ \\
& GroupLasso & ${.138_{\pm .152}}$ & ${.455_{\pm .198}}$ & ${.168_{\pm .046}}$ & ${.247_{\pm .055}}$ & ${.079_{\pm.036}}$ & ${.578_{\pm.144}}$ & ${.587_{\pm.182}}$ \\
& Improve(\%) & {\cellcolor{gray!20}{$\mathbf{+30.2}$}} & {\cellcolor{gray!20}{$\mathbf{+10.7}$}} & {\cellcolor{gray!20}{$\mathbf{+10.5}$}} & {\cellcolor{gray!20}{$\mathbf{+9.8}$}} & $-10.2$ & {\cellcolor{gray!20}{$\mathbf{+7.4}$}} & {\cellcolor{gray!20}{$\mathbf{+21.3}$}} \\
\midrule
\multirow{3}{*}{GIN}
& Lasso & $.085_{\pm.125}$ & ${.321_{\pm .248}}$ & ${.206_{\pm.092}}$ & $.205_{\pm.071}$ & ${.113_{\pm .078}}$ & ${.352_{\pm.143}}$ & $.576_{\pm.271}$ \\
& GroupLasso & ${.139_{\pm .175}}$ & ${.406_{\pm.370}}$ & ${.242_{\pm .060}}$ & ${.228_{\pm.081}}$ & ${.111_{\pm.047}}$ & ${.388_{\pm .095}}$ & ${.596_{\pm .190}}$ \\
& Improve(\%) & {\cellcolor{gray!20}{$\mathbf{+63.5}$}} & {\cellcolor{gray!20}{$\mathbf{+26.5}$}} & {\cellcolor{gray!20}{$\mathbf{+17.5}$}} & {\cellcolor{gray!20}{$\mathbf{+11.2}$}} & $-1.8$ & {\cellcolor{gray!20}{$\mathbf{+10.2}$}} & {\cellcolor{gray!20}{$\mathbf{+3.5}$}} \\
\midrule
\multirow{3}{*}{SAGE}
& Lasso & $.181_{\pm.221}$ & $.379_{\pm.281}$ & $.204_{\pm.079}$ & ${.248_{\pm .081}}$ & ${.113_{\pm.035}}$ & ${.256_{\pm .176}}$ & $.501_{\pm.248}$ \\
& GroupLasso & ${.200_{\pm .253}}$ & ${.587_{\pm .277}}$ & ${.265_{\pm .089}}$ & ${.240_{\pm.083}}$ & ${.115_{\pm .076}}$ & ${.370_{\pm.143}}$ & ${.665_{\pm .218}}$ \\
& Improve(\%) & {\cellcolor{gray!20}{$\mathbf{+10.5}$}} & {\cellcolor{gray!20}{$\mathbf{+54.9}$}} & {\cellcolor{gray!20}{$\mathbf{+29.9}$}} & $-3.2$ & {\cellcolor{gray!20}{$\mathbf{+1.8}$}} & {\cellcolor{gray!20}{$\mathbf{+44.5}$}} & {\cellcolor{gray!20}{$\mathbf{+32.7}$}} \\
\bottomrule
\end{tabular}
}
\label{tab:ablation}
\end{table}

\subsection{Main Results}

\subsubsection{Estimation performance.}
As in \cite{subgraphx}, to ensure a fair comparison, we control the sparsity of the explanation by selecting subgraphs to achieve total coverage slightly below 70\%. 
Table \ref{tab:sub_exp} presents the subgraph-level experiment results. Our method ranks first in 20 out of 28 settings (71.4\%; 4 GNN models $\times$ 7 datasets) and, notably, achieves either first or second place in \textit{all} scenarios. \textbf{These results show that our method outperforms existing baselines by effectively identifying important subgraphs}. SubgraphX shows the next best performance.

\subsubsection{Ablation study: Group Lasso vs Lasso.}
To further evaluate whether the application of the Group Lasso penalty effectively integrates substructure domain knowledge, we conduct an ablation study by replacing the Group Lasso penalty with a simple Lasso penalty, $\lambda \| \alpha \|_1$, where $\|\cdot\|_1$ denotes the $L_1$-norm. This variant, denoted as $\text{eXEL}_{\text{node}}$, no longer leverages substructure domain knowledge and computes subgraph-level importance by averaging node-level importance. 

The ablation results in Table \ref{tab:ablation} show that using the Group Lasso penalty achieves higher fidelity metric scores than using the simple Lasso penalty in almost all settings (82.1\%, 23 out of 28 scenarios). \textbf{This implies that the Group Lasso penalty effectively leverages prior domain knowledge when optimizing the score vector $\alpha$}. This supports the advantage of eXEL, where domain knowledge of graph substructures is incorporated into the sparsity penalty by grouping nodes that belong to each predefined subgraph unit.

\begin{wraptable}{r}{0.4\linewidth}
\vspace{-7mm}
\caption{
Evaluation of importance estimation against ground-truth subgraph importance. 
The best result is highlighted with \colorbox{gray!20}{color} and the second-best is \underline{underlined}.}
\setlength{\tabcolsep}{2pt}
\resizebox{\linewidth}{!}{
\begin{tabular}{lcc}
\toprule
model & PR-AUC & ROC-AUC \\
\midrule
GNNExp     & $0.574_{\pm{0.038}}$ & $0.596_{\pm{0.022}}$ \\
PGExp      & $0.404_{\pm{0.074}}$ & $0.350_{\pm{0.284}}$ \\
GradCAM    & $0.478_{\pm{0.102}}$ & $0.526_{\pm{0.237}}$ \\
SA         & $\underline{0.619_{\pm{0.107}}}$ & $\underline{0.689_{\pm{0.200}}}$ \\
GraphMask  & $0.386_{\pm{0.016}}$ & $0.524_{\pm{0.010}}$ \\
GBP        & $0.487_{\pm{0.116}}$ & $0.520_{\pm{0.106}}$ \\
\midrule
$\text{eXEL}_{\text{node}}$       & $\cellcolor{gray!20}{0.677_{\pm{0.086}}}$ & $\cellcolor{gray!20}{0.744_{\pm{0.102}}}$ \\
\bottomrule
\end{tabular}}
\label{tab:roc_auc_eval}
\vspace{-7mm}
\end{wraptable}

\subsubsection{Model analysis: Reconstruction strategy.}
Building on the previous ablation study results, we further investigate the role of the reconstruction loss through our variant, $\text{eXEL}_{\text{node}}$. When ground-truth labels are available in binary form, we treat the importance estimation task as binary classification using ROC-AUC and PR-AUC as evaluation metrics. To ensure that pretrained GNN models capture meaningful graph structures, we evaluate only those that achieve high fidelity with the ground-truth important subgraphs. 
Table \ref{tab:roc_auc_eval} reports the average scores across the datasets with ground-truth labels, where our method consistently achieves the best performance in both metrics.
This demonstrates that \textbf{even without prior domain knowledge (i.e., the Group Lasso penalty), our reconstruction strategy can \textit{partially} identify the ground-truth important subgraphs}. 

\begin{figure*}[t]
        \centering
        \begin{subfigure}[t]{0.18\linewidth}
            \includegraphics[width=\linewidth]{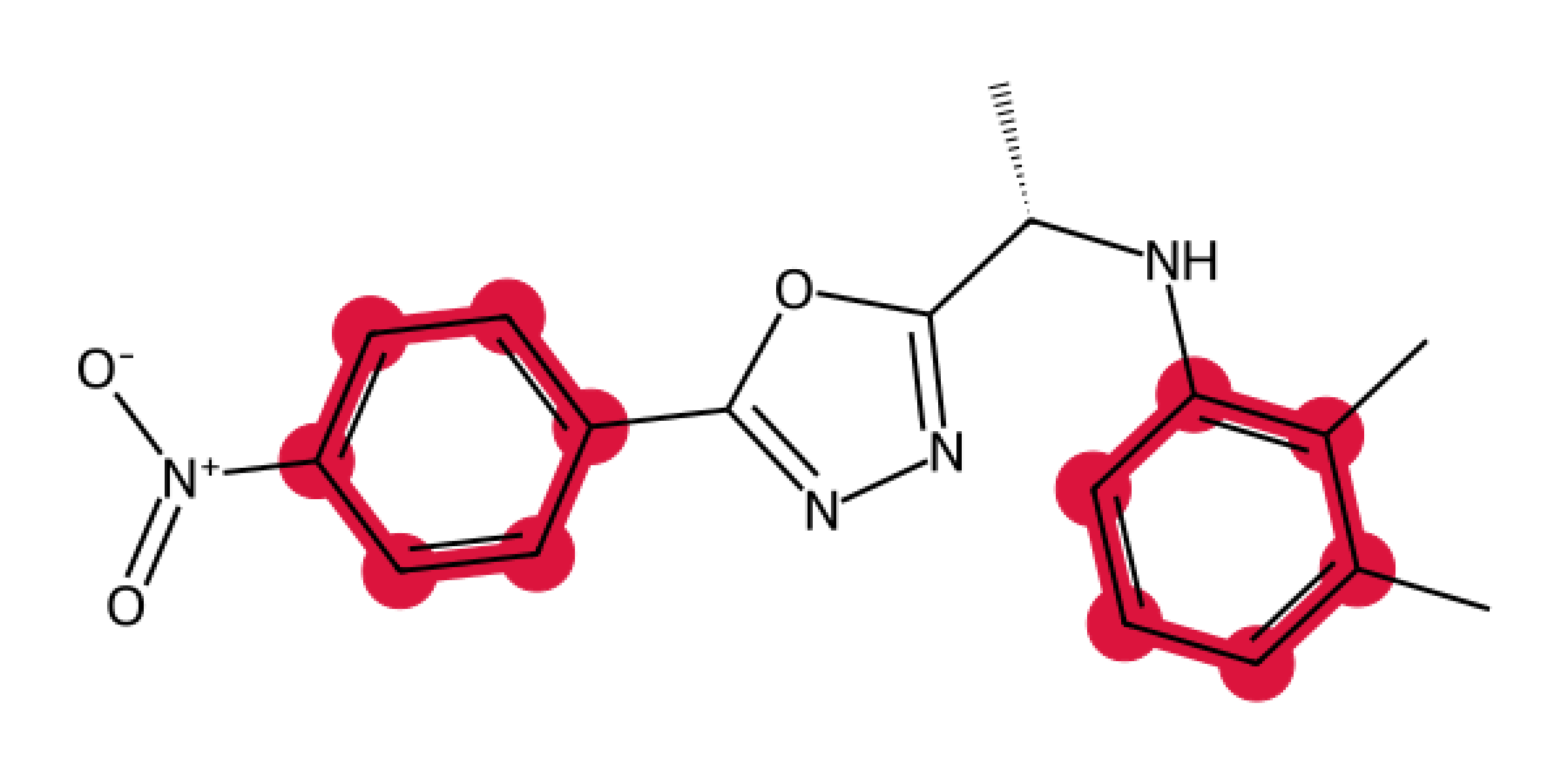}
            \caption{\scriptsize \colorbox{gray!20}{\textbf{True}}}
        \end{subfigure}
        \hfill
        \begin{subfigure}[t]{0.18\linewidth}
            \includegraphics[width=\linewidth]{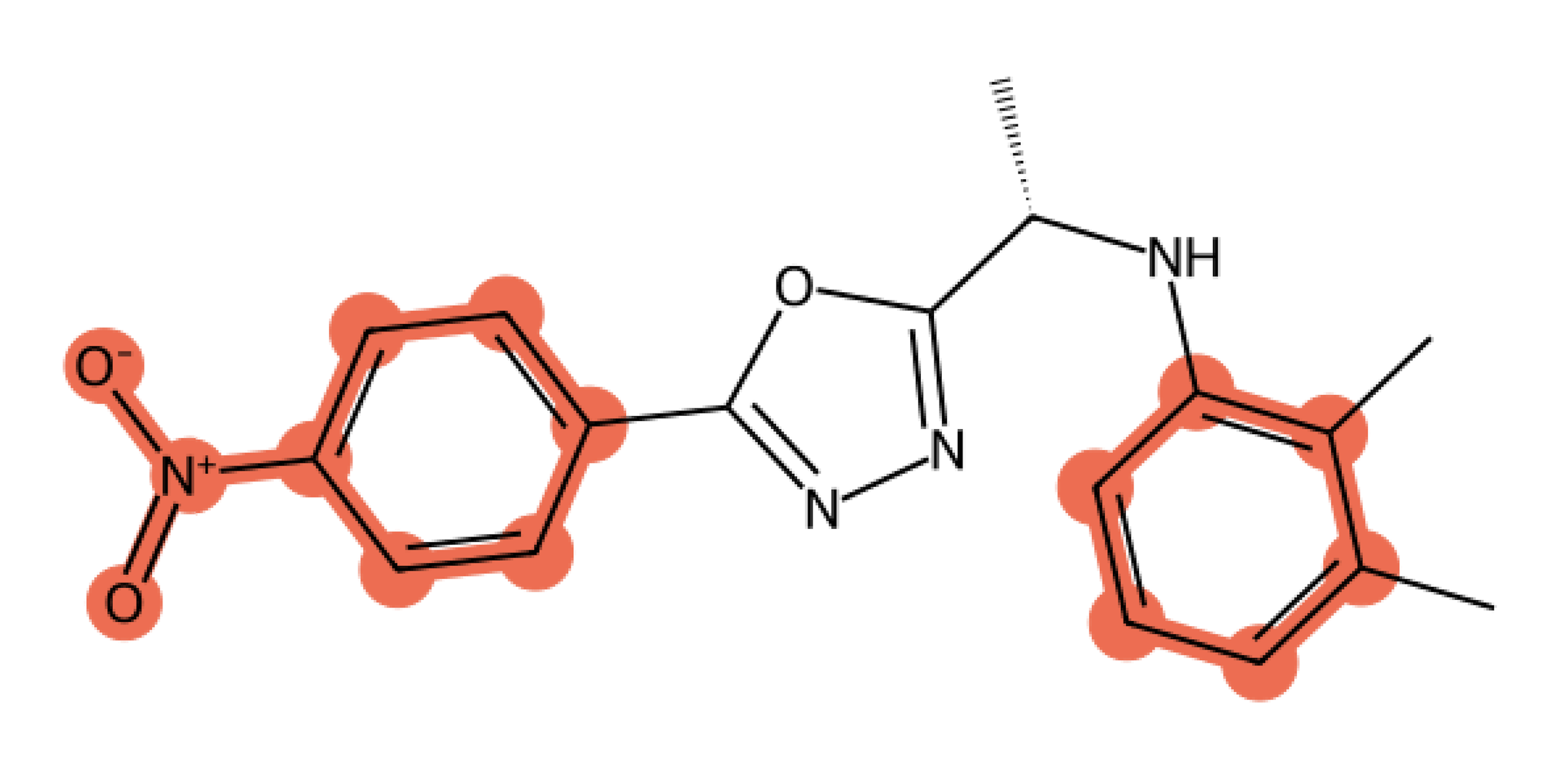}
            \caption{\scriptsize \colorbox{gray!20}{\textbf{eXEL}}}
        \end{subfigure}
        \hfill
        \begin{subfigure}[t]{0.18\linewidth}
            \includegraphics[width=\linewidth]{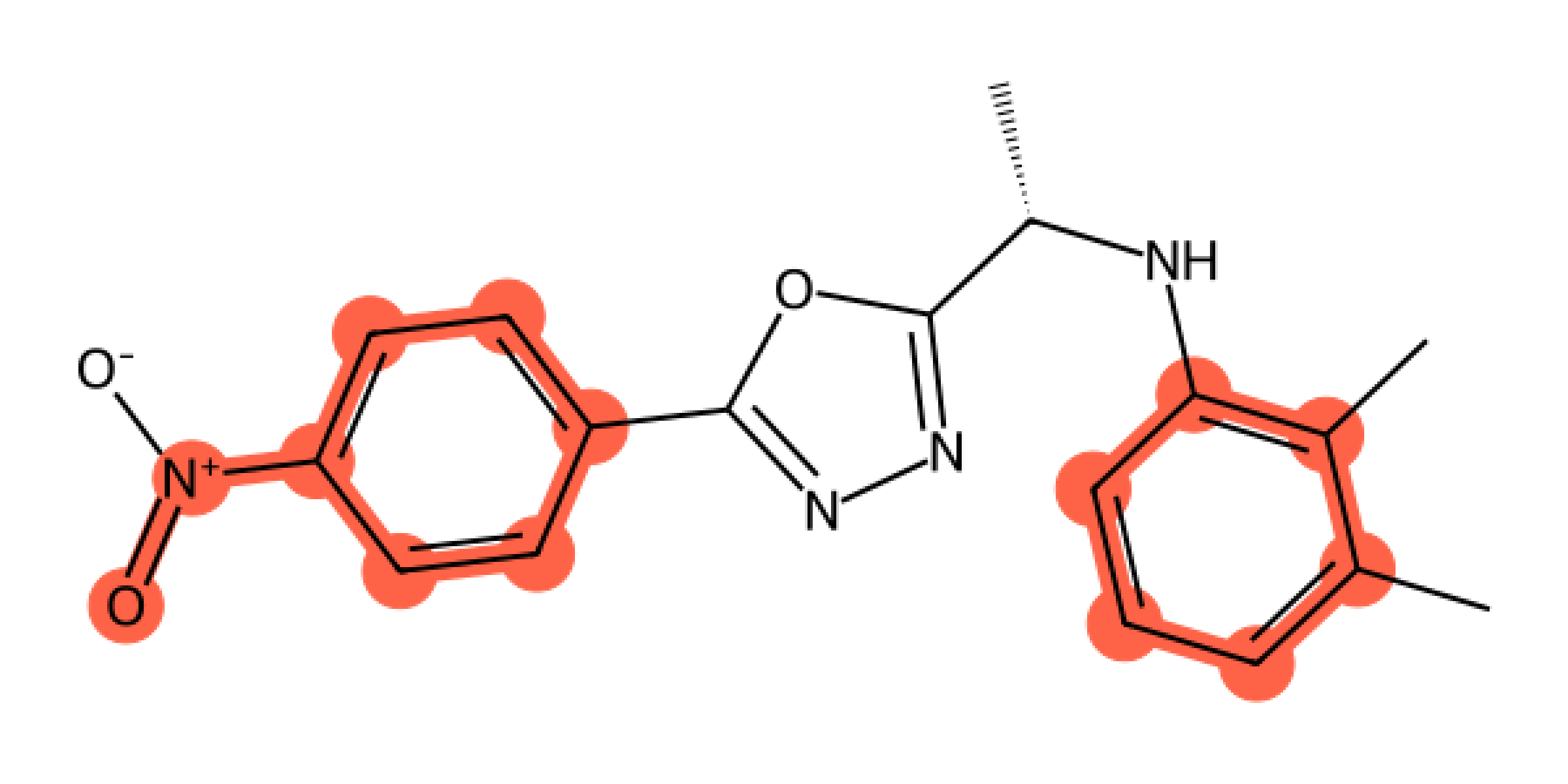}
            \caption{\scriptsize \colorbox{gray!20}{\textbf{$\text{eXEL}_{\text{node}}$}}}
        \end{subfigure}
        \hfill
        \begin{subfigure}[t]{0.18\linewidth}
            \includegraphics[width=\linewidth]{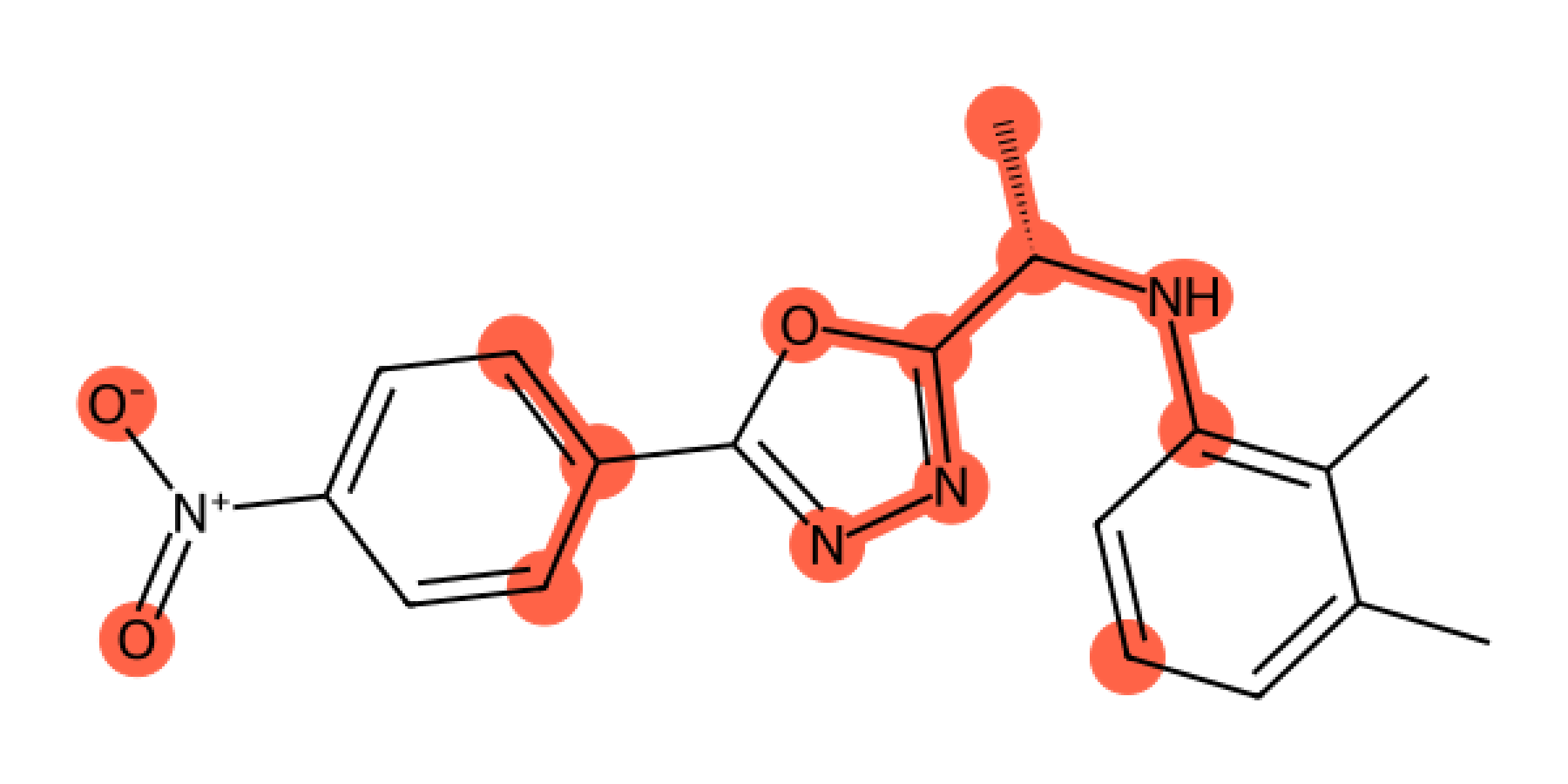}
            \caption{\scriptsize GNNExp}
        \end{subfigure}
        \hfill
        \begin{subfigure}[t]{0.18\linewidth}
            \includegraphics[width=\linewidth]{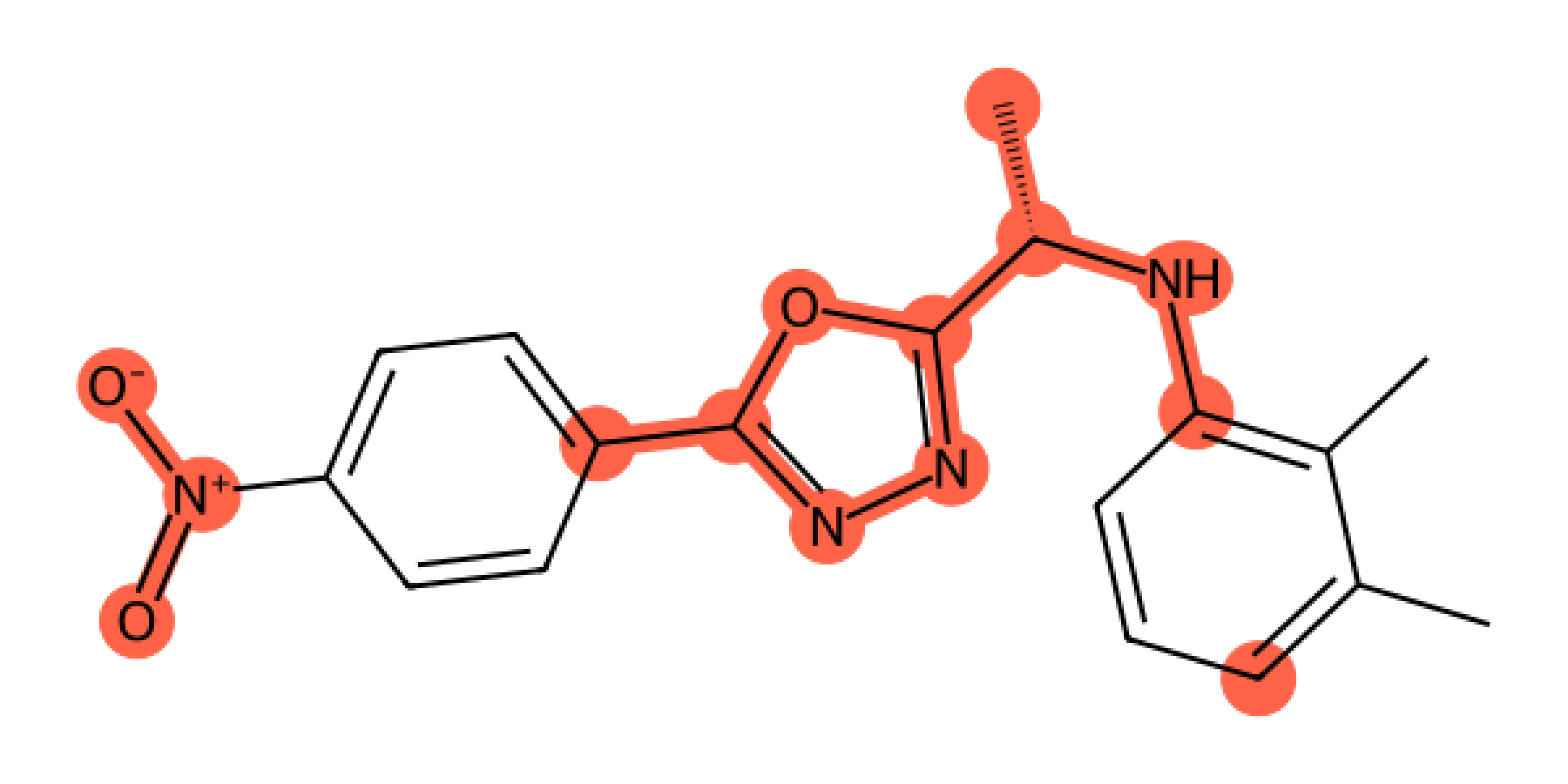}
            \caption{\scriptsize PGExp}
        \end{subfigure}
        \hfill
        \\
        \begin{subfigure}[t]{0.18\linewidth}
            \includegraphics[width=\linewidth]{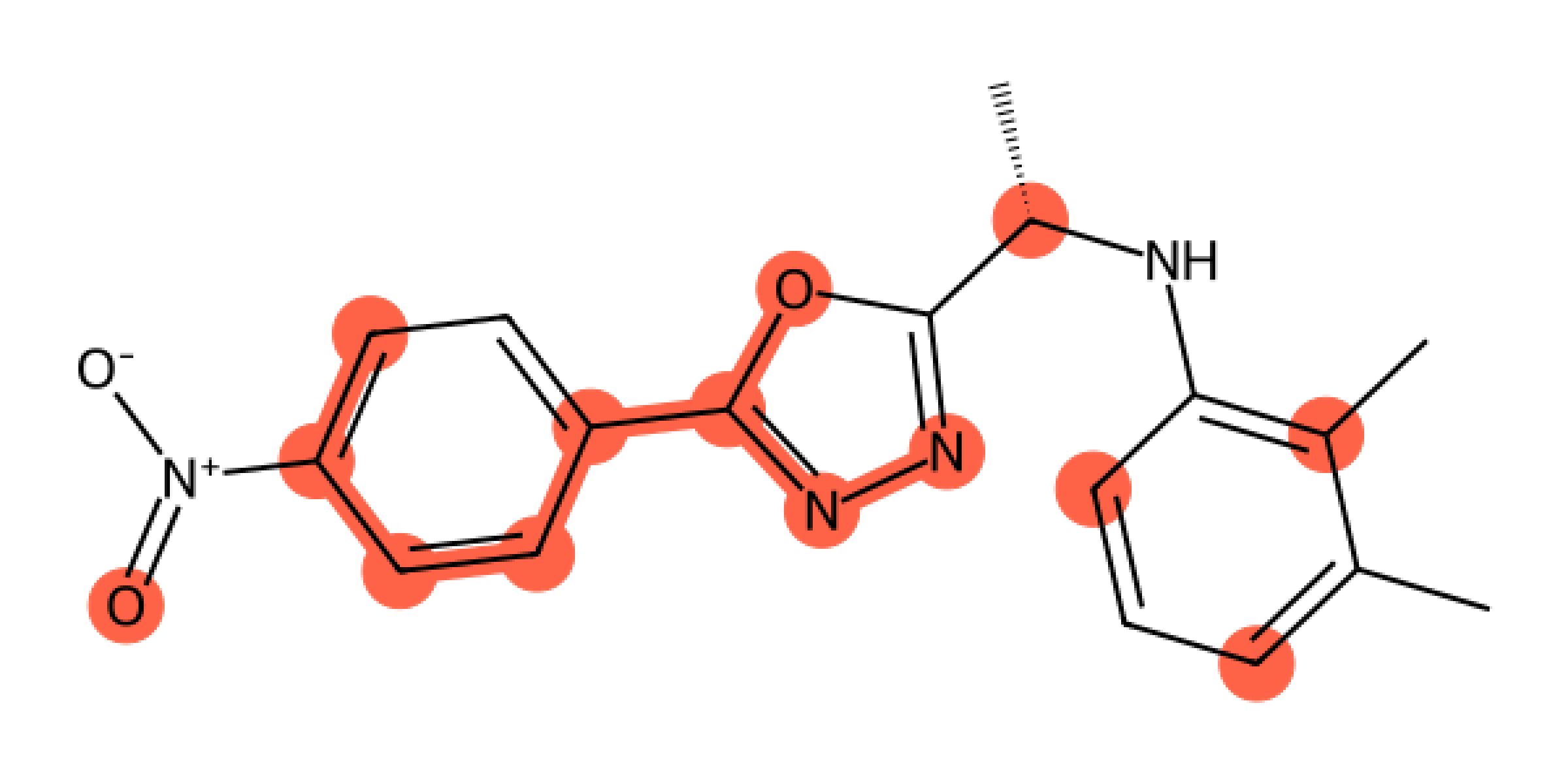}
            \caption{\scriptsize GraphMask}
        \end{subfigure}
        \hfill
        \begin{subfigure}[t]{0.18\linewidth}
            \includegraphics[width=\linewidth]{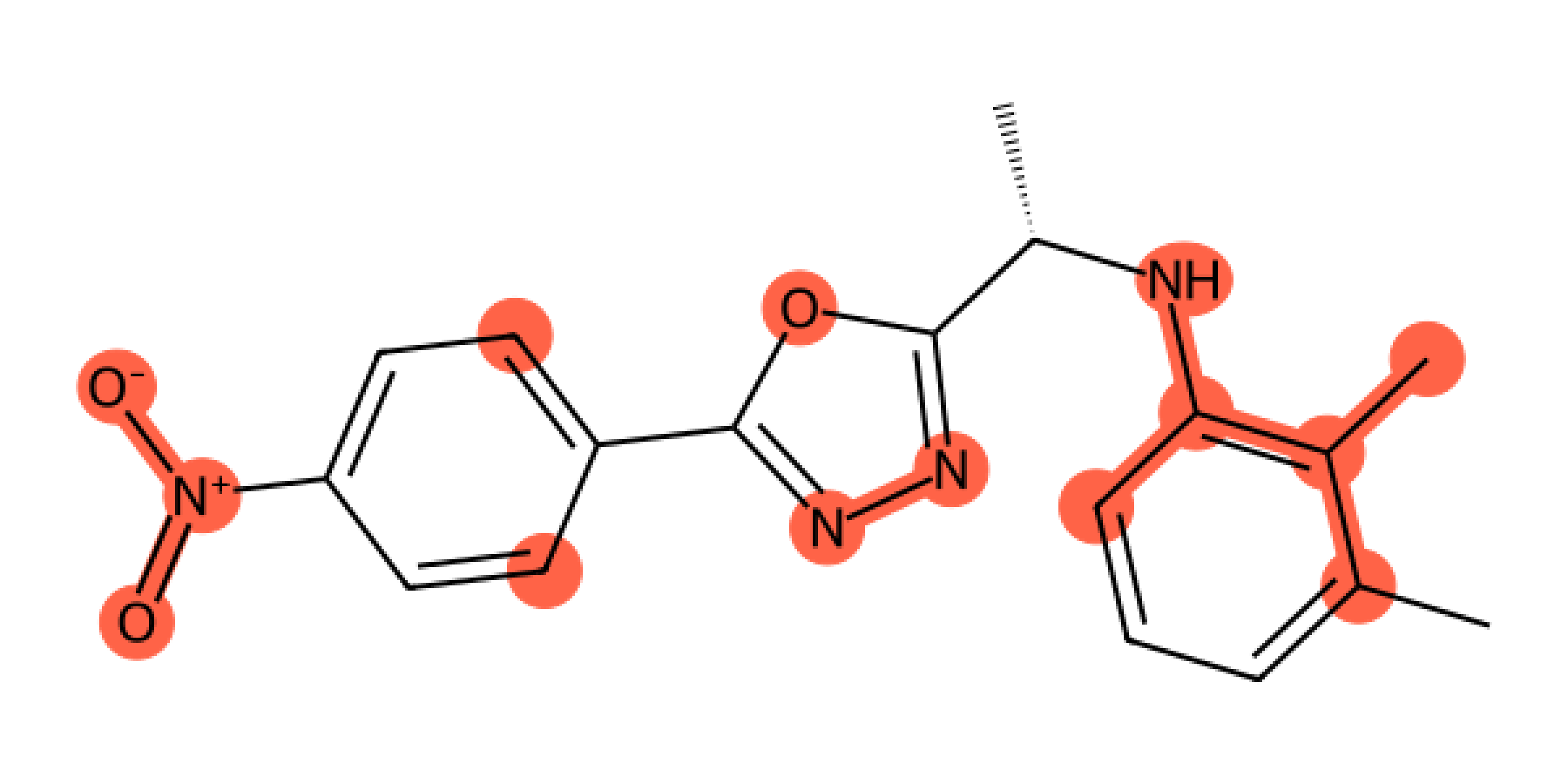}
            \caption{\scriptsize SA}
        \end{subfigure}
        \hfill
        \begin{subfigure}[t]{0.18\linewidth}
            \includegraphics[width=\linewidth]{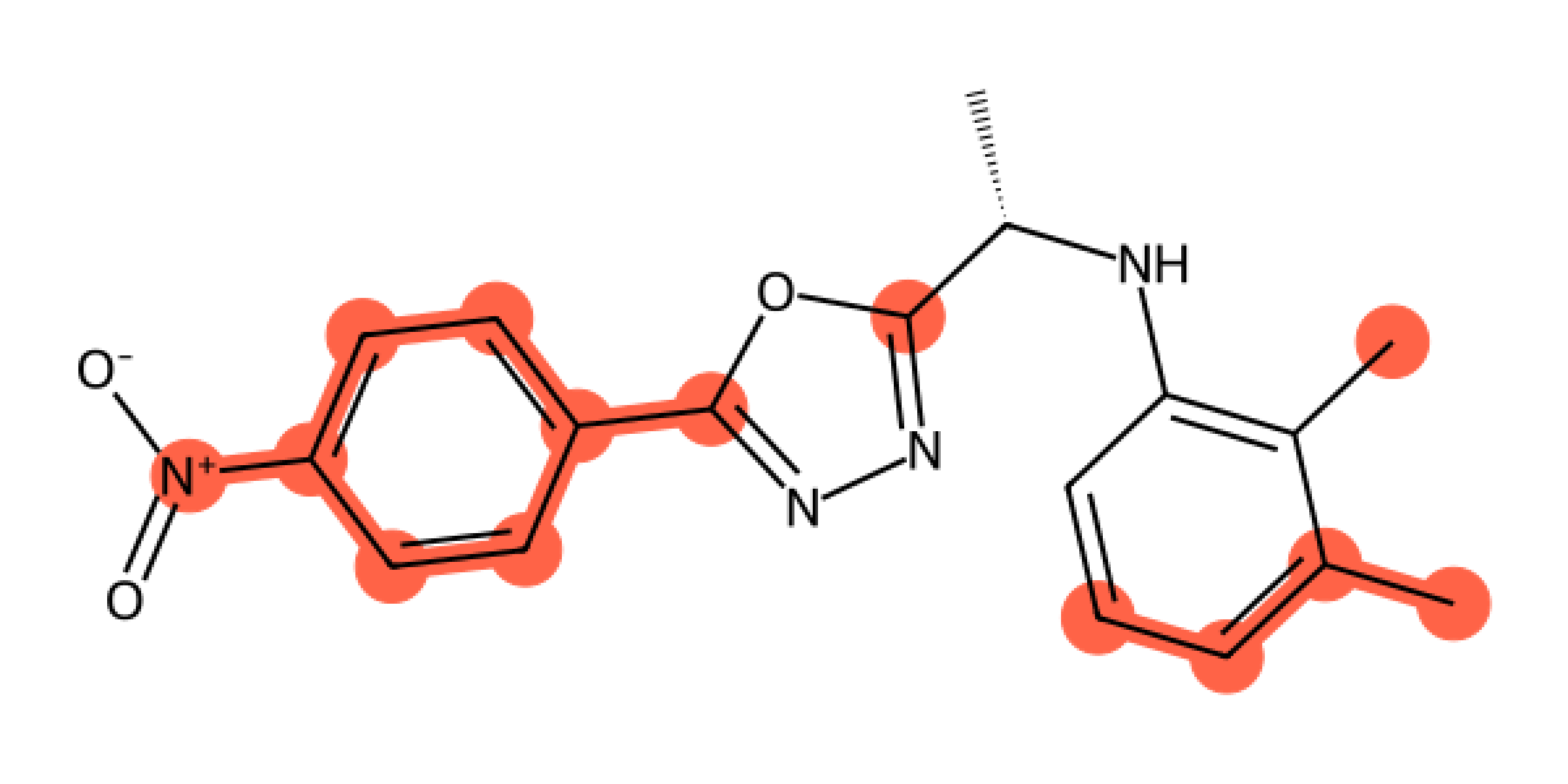}
            \caption{\scriptsize GuidedBP}
        \end{subfigure}
        \hfill
        \begin{subfigure}[t]{0.18\linewidth}
            \includegraphics[width=\linewidth]{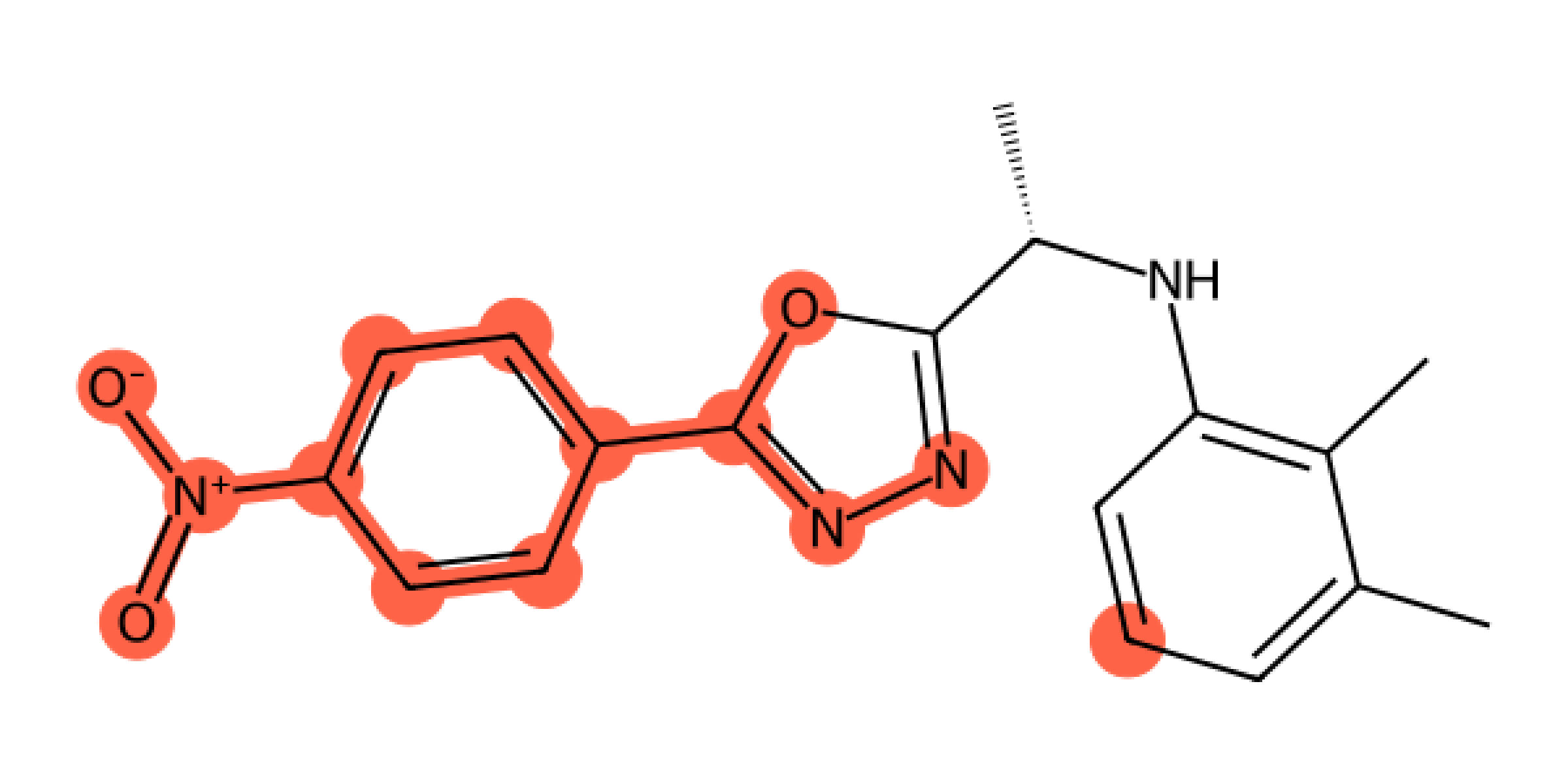}
            \caption{\scriptsize SubgraphX}
        \end{subfigure}
        \hfill
        \begin{subfigure}[t]{0.18\linewidth}
            \includegraphics[width=\linewidth]{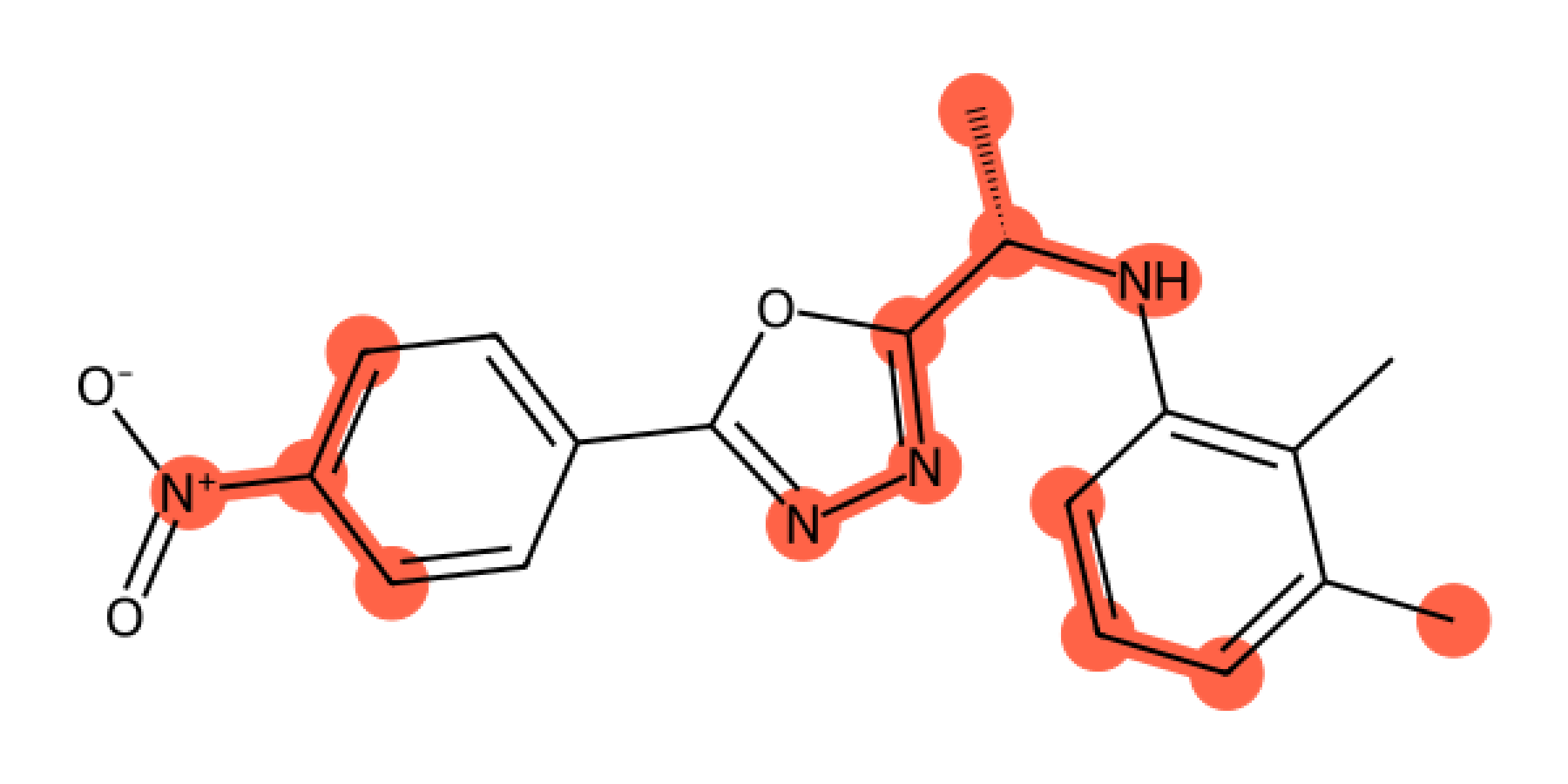}
            \caption{\scriptsize GradCAM}
        \end{subfigure}

        \begin{subfigure}[t]{0.18\linewidth}
            \includegraphics[width=\linewidth]{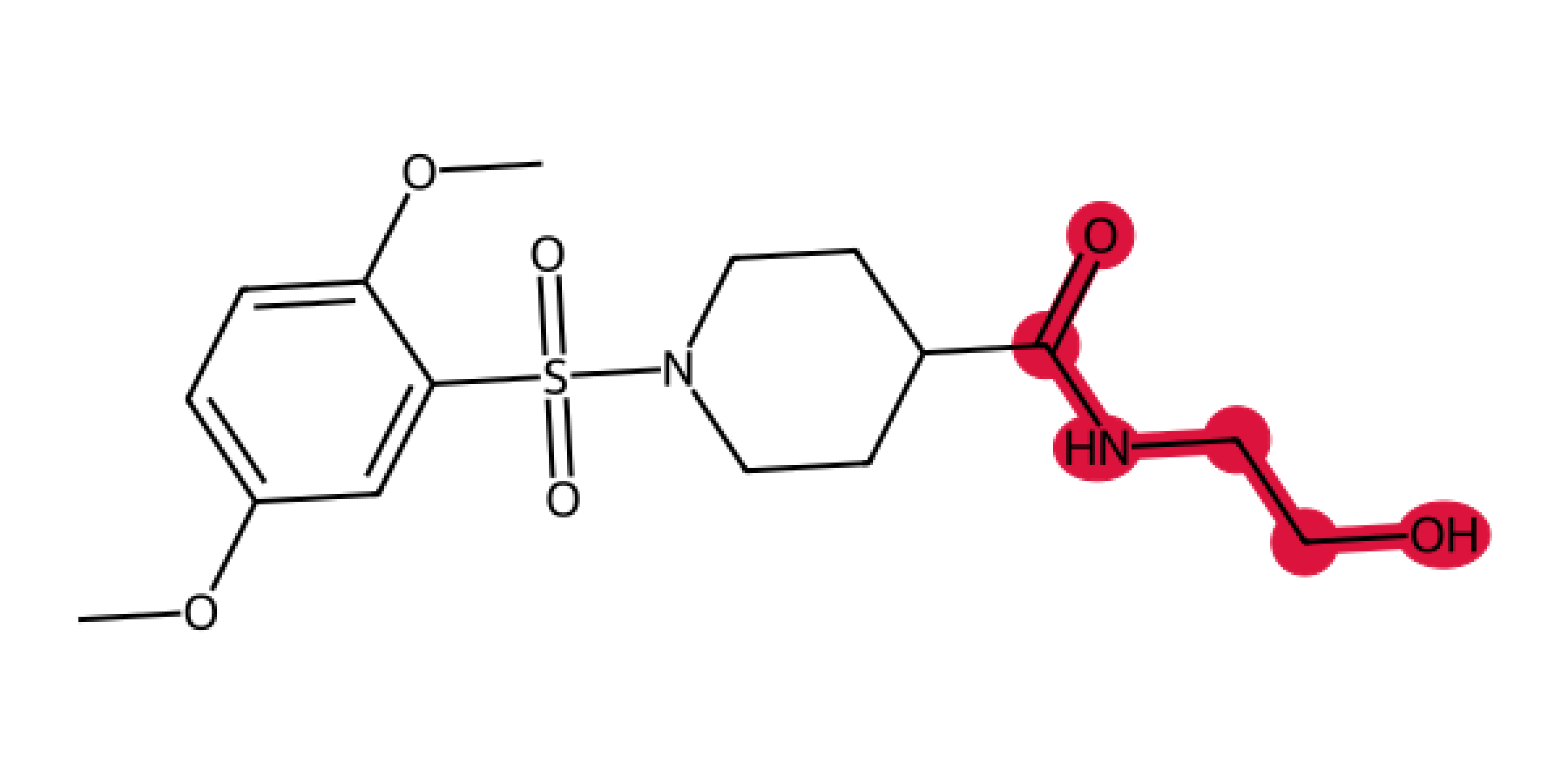}
            \caption{\scriptsize \colorbox{gray!20}{\textbf{True}}}
        \end{subfigure}
        \hfill
        \begin{subfigure}[t]{0.18\linewidth}
            \includegraphics[width=\linewidth]{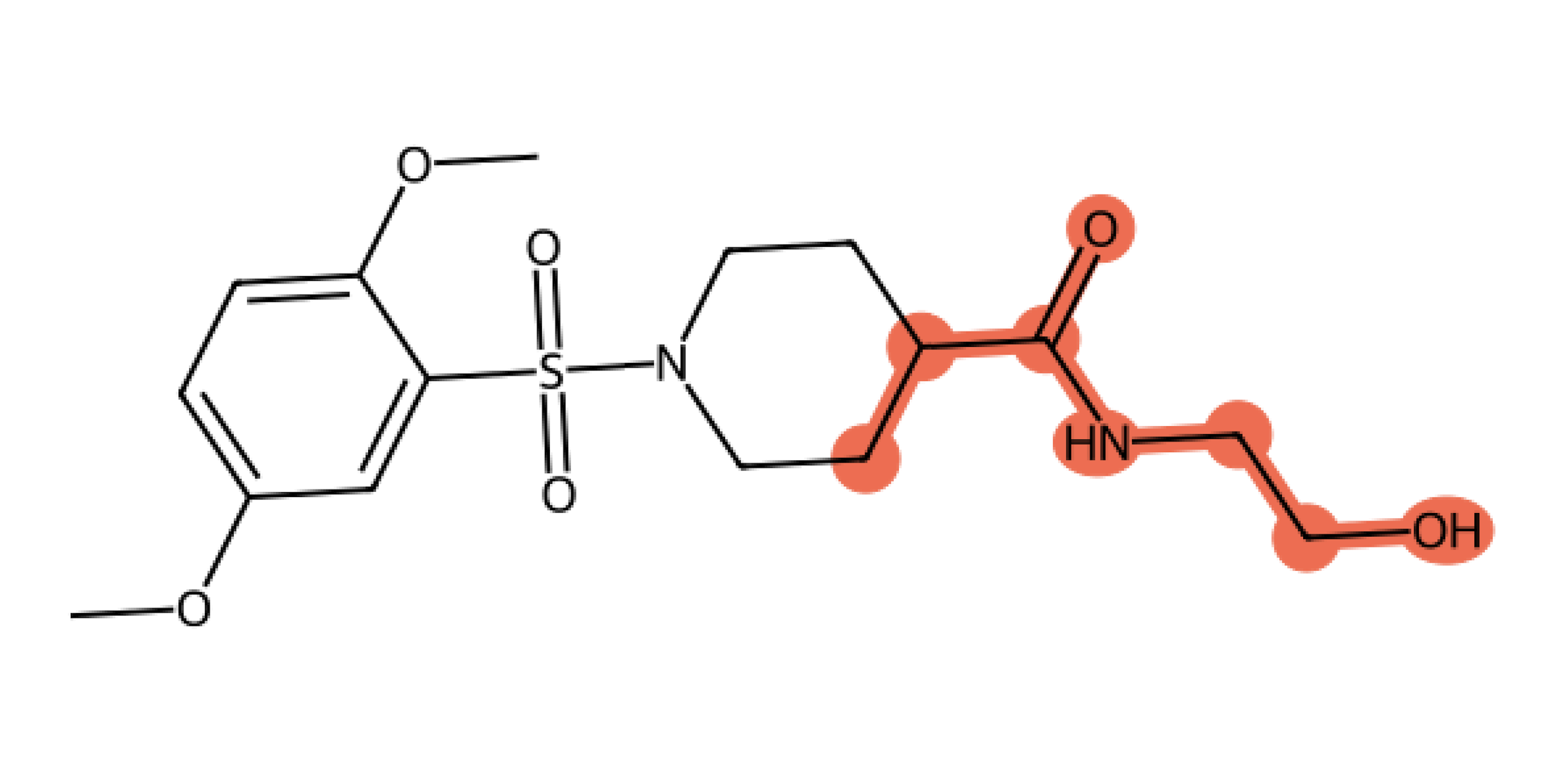}
            \caption{\scriptsize \colorbox{gray!20}{\textbf{eXEL}}}
        \end{subfigure}
        \hfill
        \begin{subfigure}[t]{0.18\linewidth}
            \includegraphics[width=\linewidth]{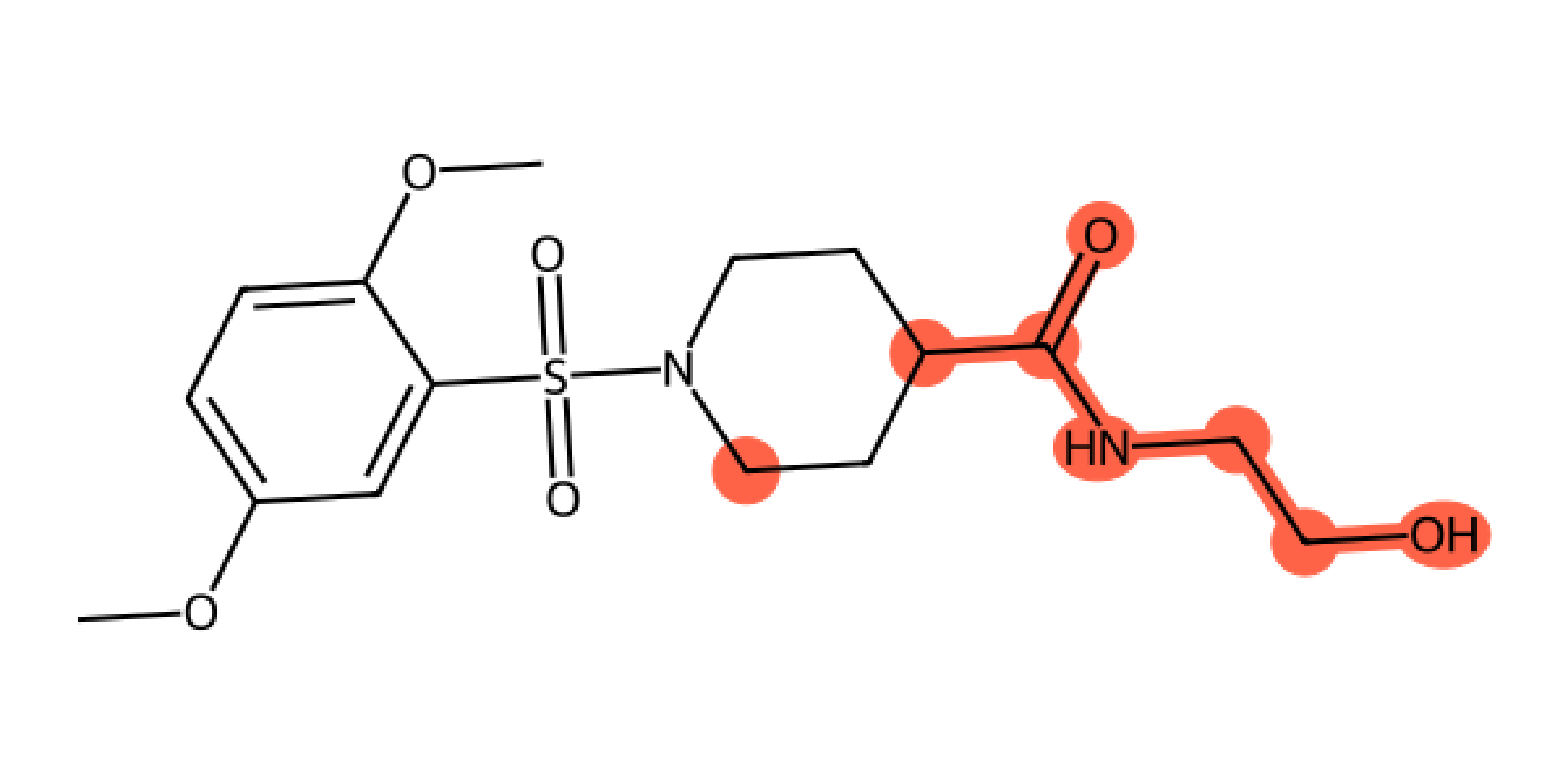}
            \caption{\scriptsize \colorbox{gray!20}{\textbf{$\text{eXEL}_{\text{node}}$}}}
        \end{subfigure}
        \hfill
        \begin{subfigure}[t]{0.18\linewidth}
            \includegraphics[width=\linewidth]{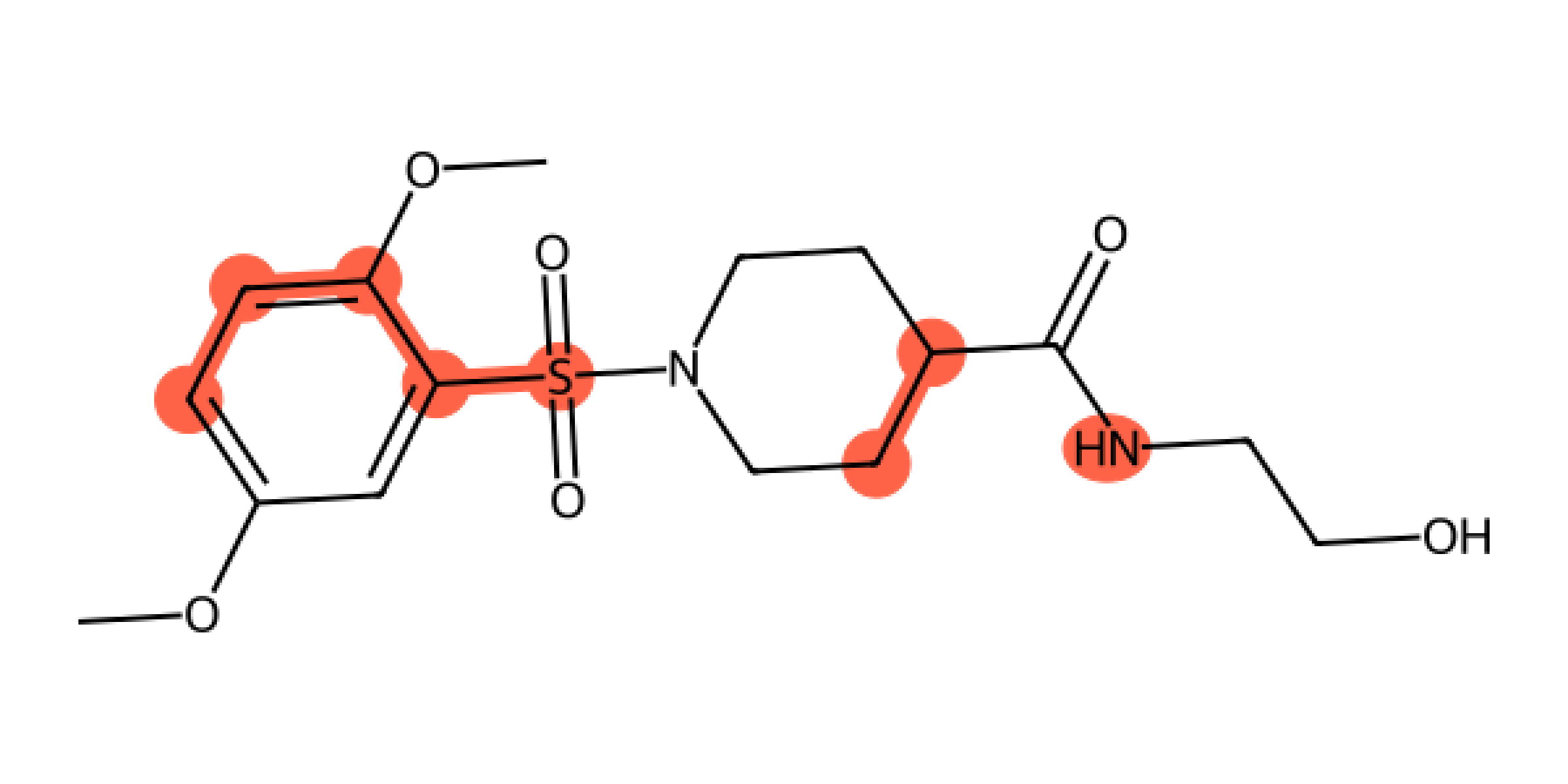}
            \caption{\scriptsize GNNExp}
        \end{subfigure}
        \hfill
        \begin{subfigure}[t]{0.18\linewidth}
            \includegraphics[width=\linewidth]{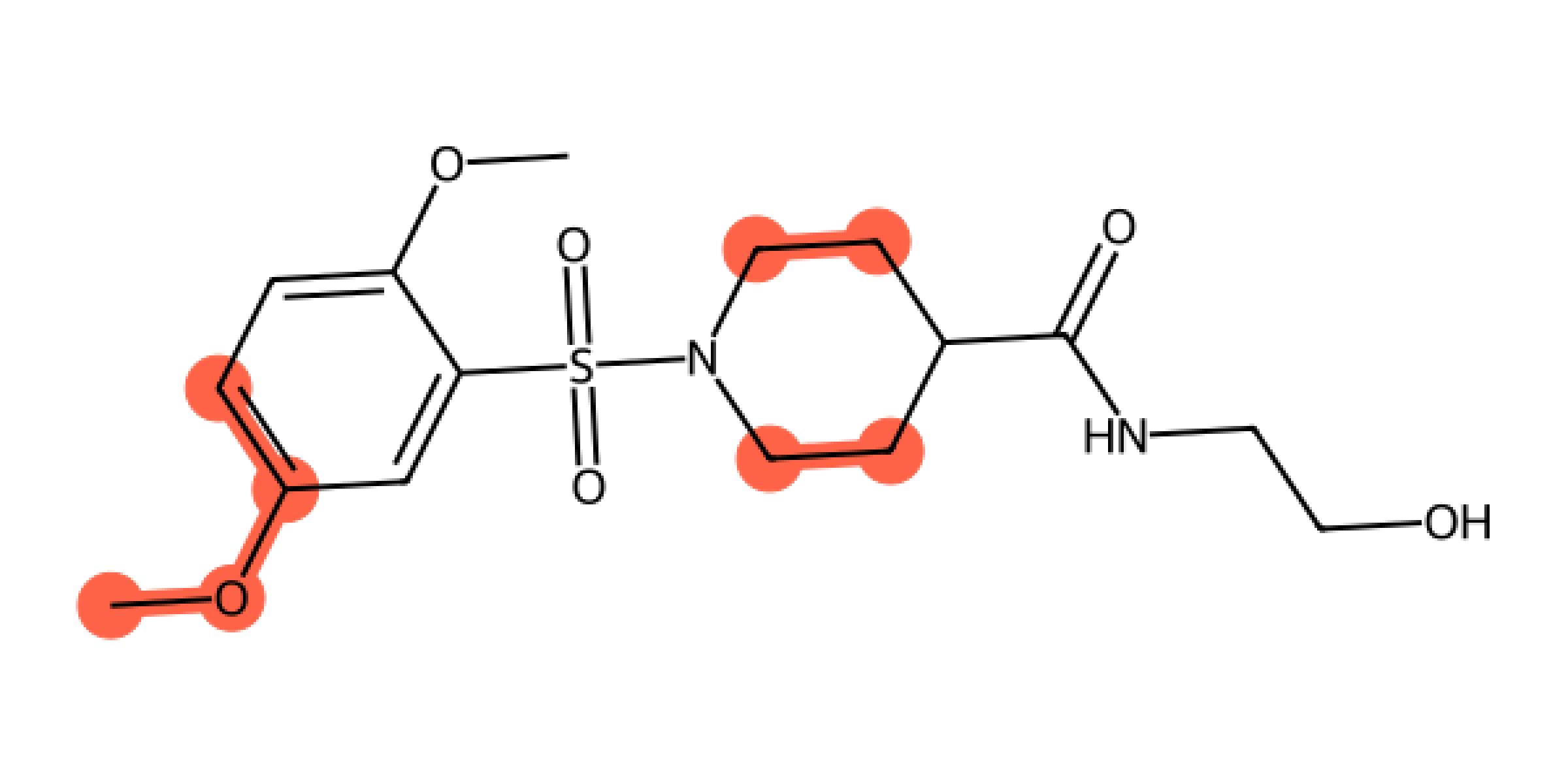}
            \caption{\scriptsize PGExp}
        \end{subfigure}
        \\
        \begin{subfigure}[t]{0.18\linewidth}
            \includegraphics[width=\linewidth]{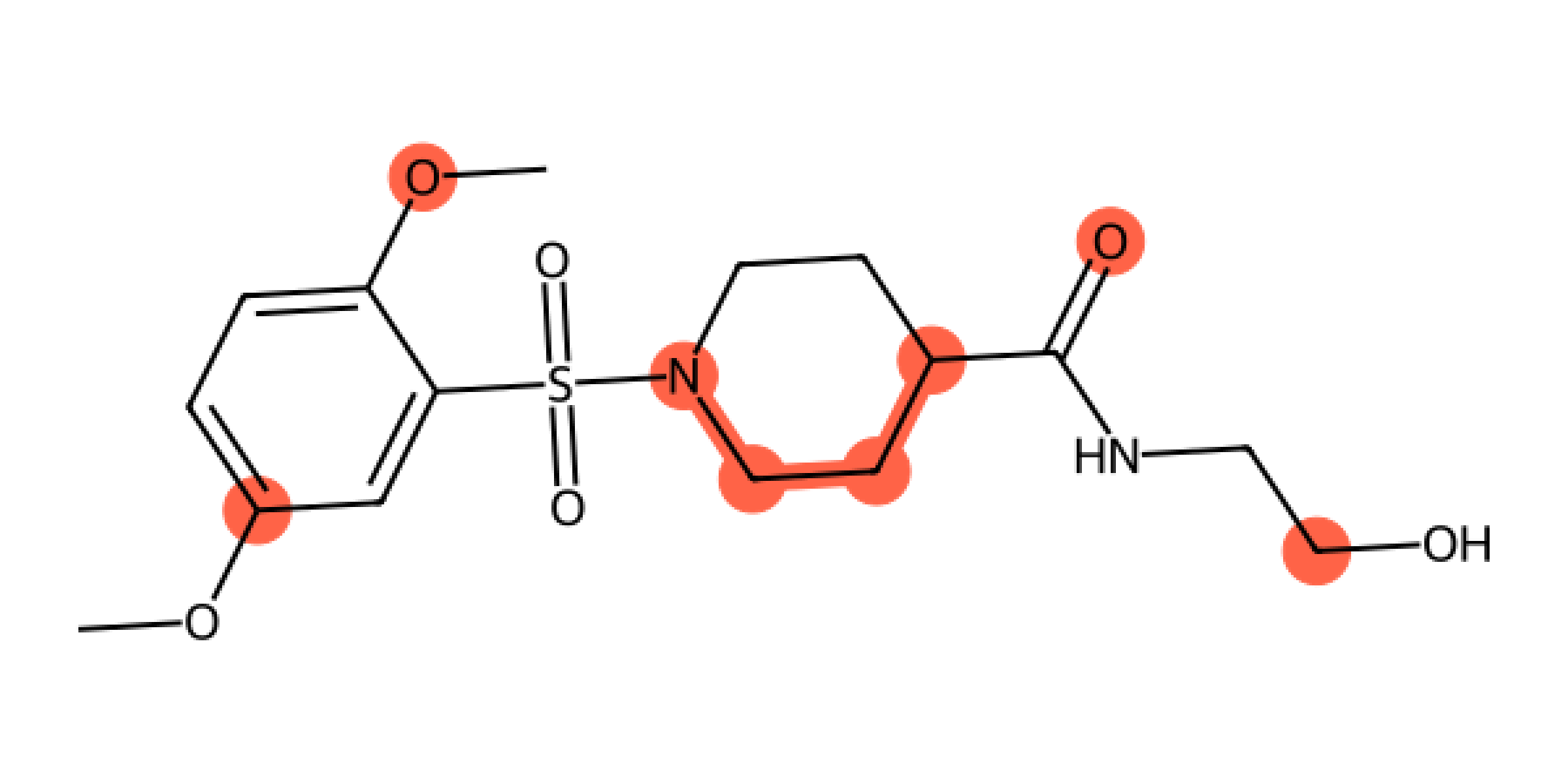}
            \caption{\scriptsize GraphMask}
        \end{subfigure}
        \hfill
        \begin{subfigure}[t]{0.18\linewidth}
            \includegraphics[width=\linewidth]{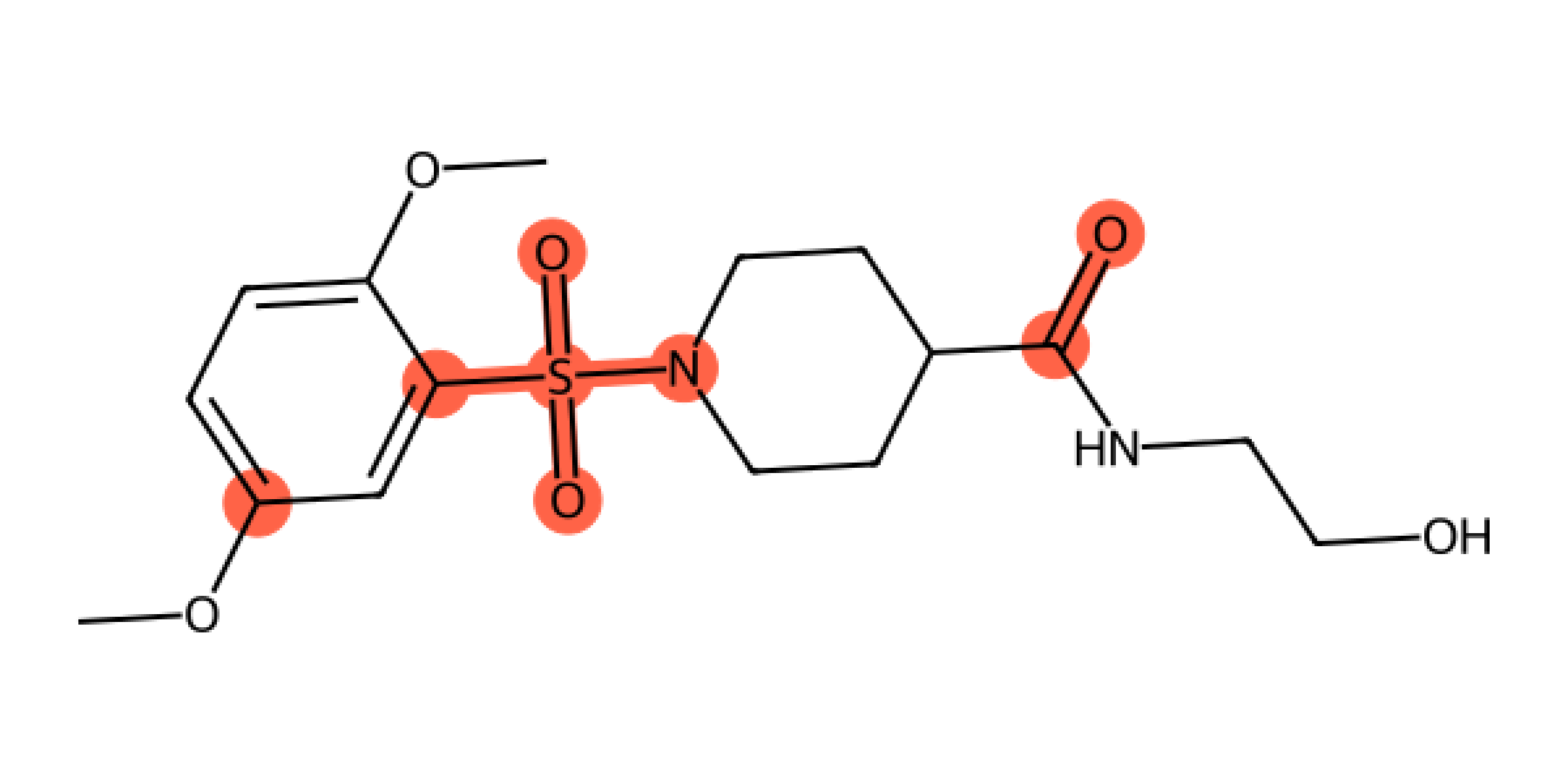}
            \caption{\scriptsize SA}
        \end{subfigure}
        \hfill
        \begin{subfigure}[t]{0.18\linewidth}
            \includegraphics[width=\linewidth]{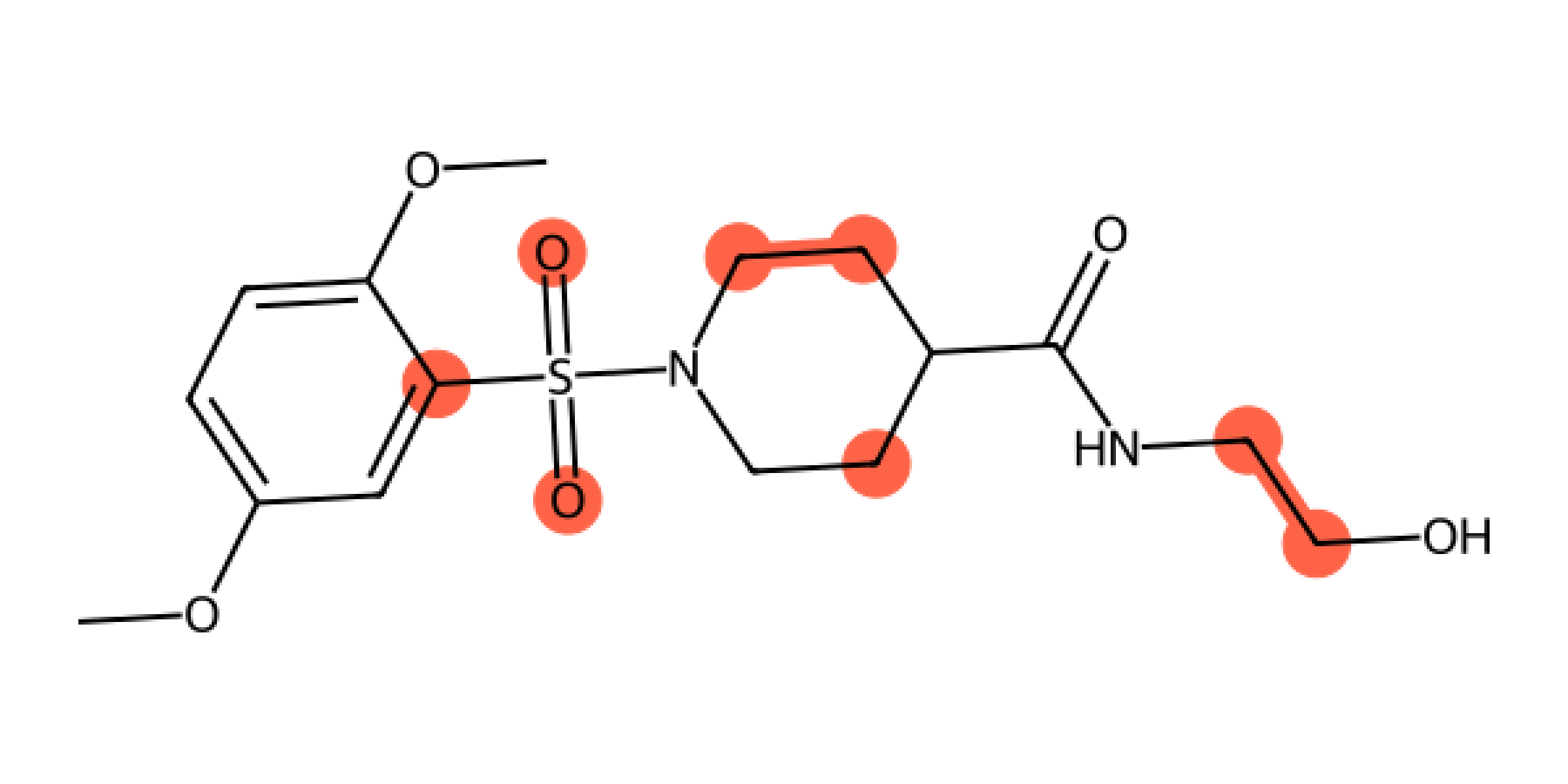}
            \caption{\scriptsize GuidedBP}
        \end{subfigure}
        \hfill
        \begin{subfigure}[t]{0.18\linewidth}
            \includegraphics[width=\linewidth]{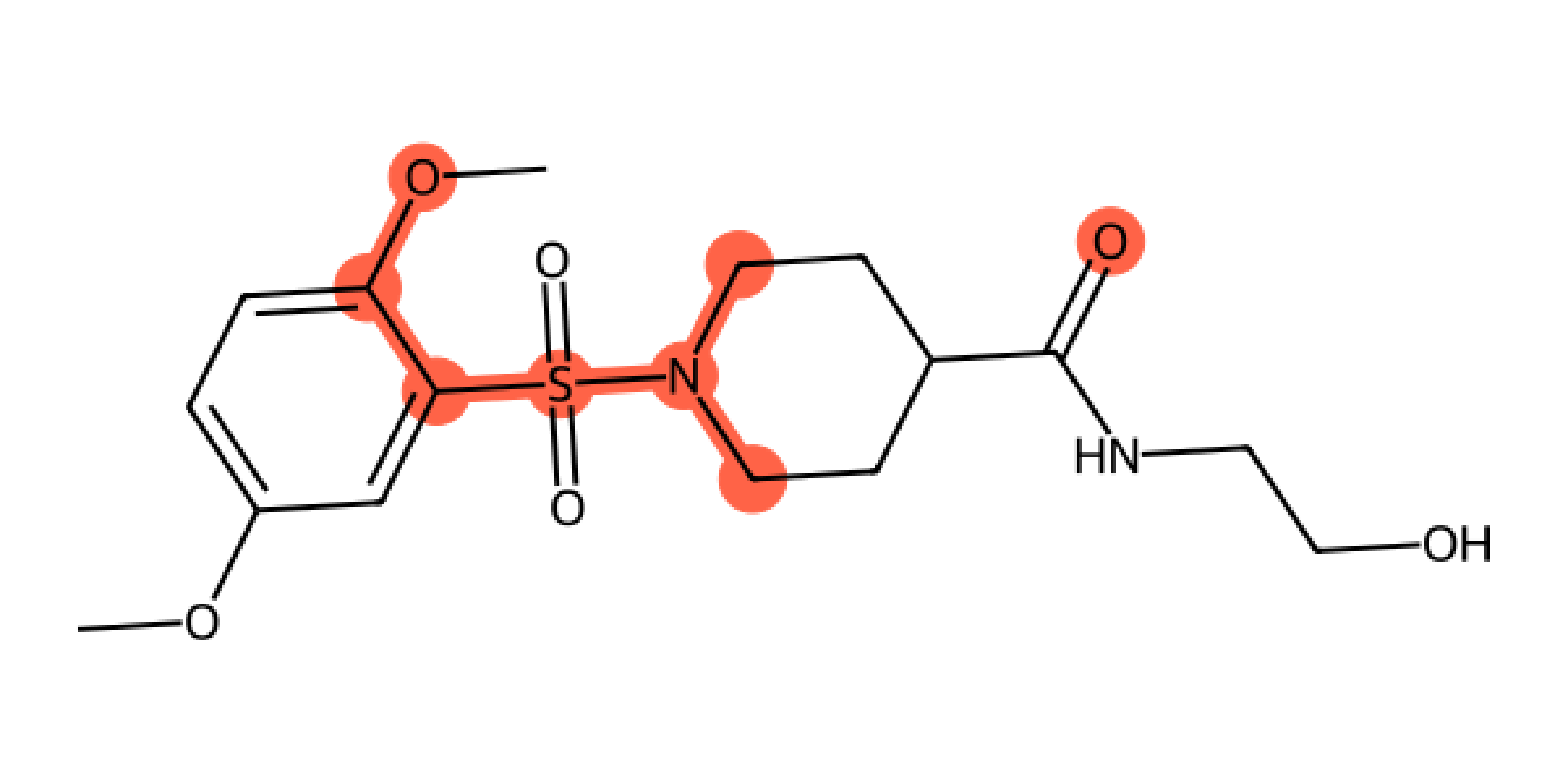}
            \caption{\scriptsize SubgraphX}
        \end{subfigure}
        \hfill
        \begin{subfigure}[t]{0.18\linewidth}
            \includegraphics[width=\linewidth]{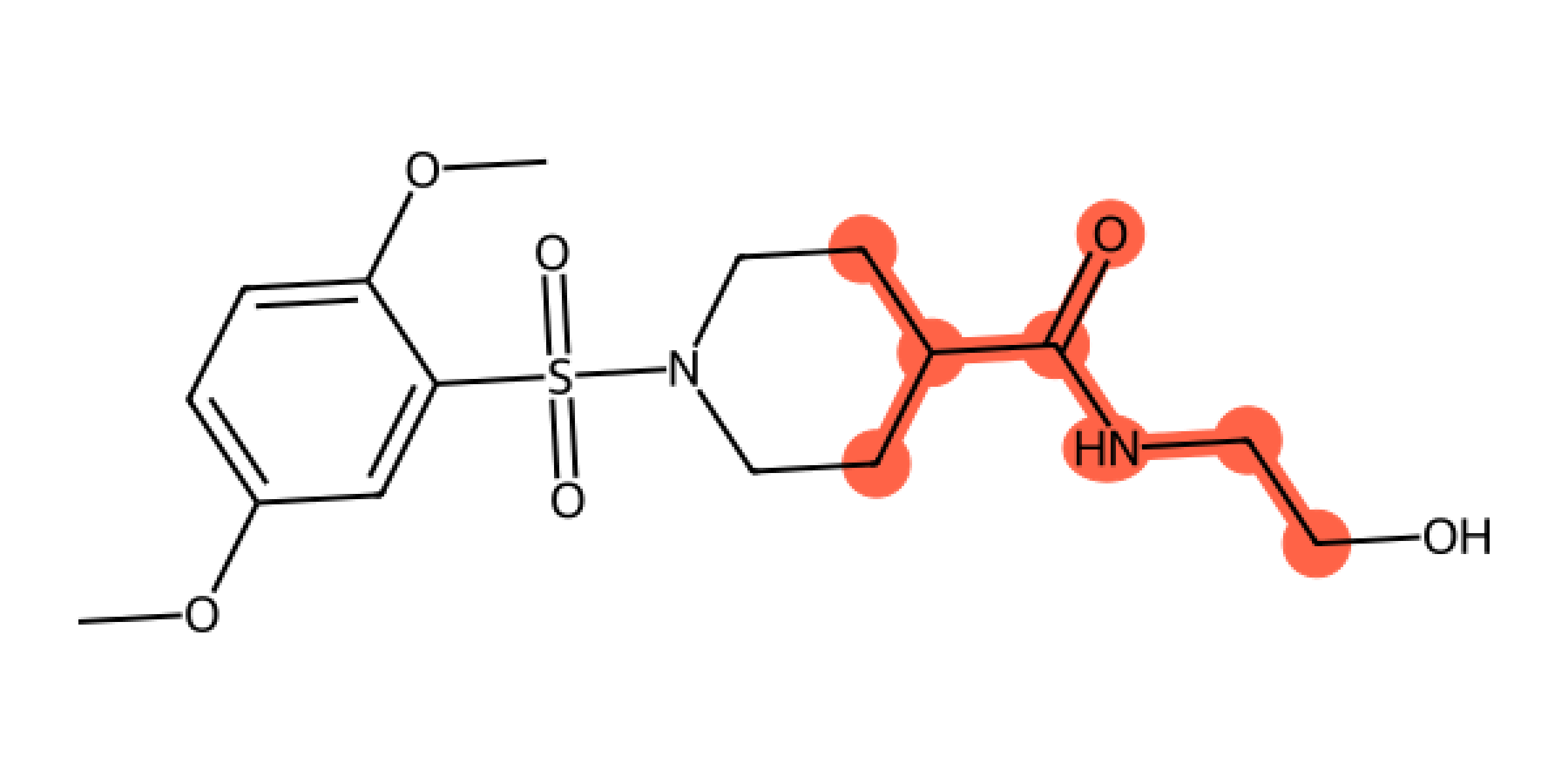}
            \caption{\scriptsize GradCAM}
        \end{subfigure}

    \caption{
    Identified important node subsets (top two rows: \texttt{Benzene}; bottom two rows: \texttt{Fluoride}). 
    The tasks are to determine whether a molecule contains a benzene ring, a fluoride group, or a carbonyl group.
    }
    \label{fig:gt_visual}
\end{figure*}

However, this partial identification occurs because the reconstruction strategy alone cannot identify \textit{connected} subgraph-level structures, which domain experts consider semantically valid.
We visualize the identified important node subsets in Figure \ref{fig:gt_visual}. As shown, $\text{eXEL}_{\text{node}}$, unlike eXEL, fails to identify important node subsets that form connected subgraph units. 
This implies that the Group Lasso penalty reinforces the reconstruction strategy by ensuring the identification of semantically valid subgraphs.

\begin{table}[t]
\scriptsize
\centering
\caption{
Node-level $\texttt{Fidelity}_{\mathrm{F_1}}$ scores. The best result is highlighted with \colorbox{gray!20}{color} and the second-best is \underline{underlined}. Higher is better. 
}
\setlength{\tabcolsep}{0.5mm}
\resizebox{0.99\textwidth}{!}{
\begin{tabular}{ll|c|cccccccc}
\toprule
model & data & $\text{eXEL}_{\text{node}}$ & GNNExp & GradCAM & GraphMask & GBP & SA & SubgraphX & PGExp \\
\midrule
\multirow{6}{*}{GCN}
 & \texttt{ENZYME} & $\cellcolor{gray!20}{.150_{\pm.079}}$ & $.083_{\pm.078}$ & $.113_{\pm.096}$ & $.049_{\pm.063}$ &  $.109_{\pm.097}$ & $\underline{.147_{\pm.075}}$ &$.072_{\pm.113}$ & $.066_{\pm.110}$ \\
 & \texttt{IMDB}   & $\underline{.021_{\pm.055}}$ & $\cellcolor{gray!20}{.024_{\pm.034}}$ & $.008_{\pm.039}$ & $.010_{\pm.033}$ & $.006_{\pm.046}$ & $.000_{\pm.049}$ & $.012_{\pm.039}$ & $.013_{\pm.034}$ \\
 & \texttt{PROTE} & $\cellcolor{gray!20}{.103_{\pm.089}}$ & $.061_{\pm.057}$ & $.059_{\pm.053}$ & $.061_{\pm.057}$ &   $.055_{\pm.066}$ & $.054_{\pm.069}$ & $\underline{.089_{\pm.077}}$ & $.007_{\pm.051}$ \\
 & \texttt{UPFD}   & $\cellcolor{gray!20}{.100_{\pm.052}}$ & $.007_{\pm.023}$ & $\underline{.032_{\pm.026}}$ & $.019_{\pm.020}$ & $.024_{\pm.024}$ & $.010_{\pm.018}$ & $.012_{\pm.031}$ & $.002_{\pm.013}$ \\
  & \texttt{MOTIF}  & $\underline{.659_{\pm.232}}$ & $.226_{\pm.097}$ & $.243_{\pm.156}$ & $.244_{\pm.142}$ & $.490_{\pm.328}$ & $.363_{\pm.325}$ & $\cellcolor{gray!20}{.791_{\pm.212}}$ & ${.429_{\pm.360}}$ \\
 & \texttt{MULTI}  & $\underline{.695_{\pm.219}}$ & $.255_{\pm.107}$ & $.237_{\pm.136}$ & $.269_{\pm.156}$ & $.452_{\pm.286}$ & $.355_{\pm.335}$ & $\cellcolor{gray!20}{.864_{\pm.157}}$ & ${.466_{\pm.353}}$ \\
\midrule

\multirow{6}{*}{GAT}
 & \texttt{ENZYME} & $\cellcolor{gray!20}{.180_{\pm.085}}$ & $.138_{\pm.086}$ & $.126_{\pm.080}$ & $.151_{\pm.050}$ &  $.151_{\pm.076}$ & $\underline{.162_{\pm.085}}$ & $.124_{\pm.067}$ & $.144_{\pm.077}$ \\
 & \texttt{IMDB}    & $\cellcolor{gray!20}{.061_{\pm.043}}$ & $.039_{\pm.037}$ & $.029_{\pm.038}$ & $.033_{\pm.045}$ &   $.020_{\pm.057}$ & $\underline{.043_{\pm.058}}$ & $.032_{\pm.045}$ & $.030_{\pm.044}$ \\
 & \texttt{PROTE} & $\cellcolor{gray!20}{.104_{\pm.074}}$ & $.078_{\pm.069}$ & $.045_{\pm.044}$ & $.077_{\pm.076}$ & $.077_{\pm.070}$ & $.064_{\pm.065}$ & $\underline{.092_{\pm.105}}$ & ${.084_{\pm.080}}$ \\
 & \texttt{UPFD}   & $\cellcolor{gray!20}{.146_{\pm.082}}$ & $.034_{\pm.026}$ & $\underline{.086_{\pm.078}}$ & $.020_{\pm.024}$ & $.030_{\pm.017}$ & $.077_{\pm.070}$ & $.019_{\pm.036}$ & $.018_{\pm.031}$ \\
  & \texttt{MOTIF} & $\underline{.312_{\pm.137}}$ & $.203_{\pm.070}$ & $.110_{\pm.182}$ & $.170_{\pm.052}$ & $.143_{\pm.136}$ & $.162_{\pm.161}$ & $\cellcolor{gray!20}{.406_{\pm.226}}$ & ${.164_{\pm.237}}$ \\
 & \texttt{MULTI}  & $\underline{.309_{\pm.069}}$ & $.207_{\pm.067}$ & $.131_{\pm.186}$ & $.156_{\pm.053}$ & $.144_{\pm.139}$ & $.206_{\pm.213}$ & $\cellcolor{gray!20}{.424_{\pm.239}}$ & ${.149_{\pm.239}}$ \\
\midrule

\multirow{6}{*}{GIN}
 & \texttt{ENZYME} & $\underline{.176_{\pm.076}}$ & $.091_{\pm.093}$ & $\cellcolor{gray!20}{.177_{\pm.094}}$ & $.093_{\pm.077}$ & $.174_{\pm.086}$ & $.153_{\pm.067}$ & $.116_{\pm.091}$ & $.108_{\pm.108}$ \\
 & \texttt{IMDB}   & $.022_{\pm.055}$ & $.014_{\pm.018}$ & $.008_{\pm.047}$ & $.003_{\pm.026}$ & $.018_{\pm.041}$ & $.019_{\pm.046}$ & $\underline{.023_{\pm.048}}$ & $\cellcolor{gray!20}{.028_{\pm.023}}$ \\

 & \texttt{PROTE} & $\underline{.035_{\pm.041}}$ & $.017_{\pm.045}$ & $.010_{\pm.043}$ & $.009_{\pm.051}$ &   $.020_{\pm.038}$ & $\cellcolor{gray!20}{.042_{\pm.082}}$ & $.028_{\pm.050}$ & $.000_{\pm.037}$ \\
 & \texttt{UPFD}   & $\cellcolor{gray!20}{.129_{\pm.095}}$ & $.005_{\pm.031}$ & $\underline{.149_{\pm.106}}$ & $.000_{\pm.016}$ & $.029_{\pm.035}$ & $.005_{\pm.036}$ & $.006_{\pm.029}$ & $.001_{\pm.020}$ \\

  & \texttt{MOTIF}  & $\cellcolor{gray!20}{.800_{\pm.125}}$ & $.164_{\pm.056}$ & $.564_{\pm.245}$ & $.130_{\pm.029}$ & ${.587_{\pm.391}}$ & $.539_{\pm.297}$ & $\underline{.734_{\pm.069}}$ & $.006_{\pm.010}$ \\
 & \texttt{MULTI}  & $\cellcolor{gray!20}{.793_{\pm.137}}$ & $.166_{\pm.071}$ & $.541_{\pm.268}$ & $.138_{\pm.039}$ & ${.596_{\pm.401}}$ & $.456_{\pm.297}$ & $\underline{.731_{\pm.082}}$ & $.004_{\pm.008}$ \\
\midrule
\multirow{6}{*}{SAGE}
 & \texttt{ENZYME} & $.117_{\pm.090}$ & $.101_{\pm.091}$ & $.112_{\pm.115}$ & $.078_{\pm.095}$ & $\cellcolor{gray!20}{.183_{\pm.073}}$ & $\underline{.132_{\pm.088}}$ & $.076_{\pm.076}$ & $.100_{\pm.087}$ \\
 &\texttt{IMDB}  & $\underline{.044_{\pm.068}}$ & \cellcolor{gray!20}$.054_{\pm.079}$ & $.009_{\pm.073}$ & ${.043_{\pm.069}}$ & $.042_{\pm.068}$ & $.027_{\pm.068}$ & $.001_{\pm.071}$ & $.042_{\pm.085}$ \\

 & \texttt{PROTE} & $\cellcolor{gray!20}{.095_{\pm.043}}$ & $\underline{.092_{\pm.083}}$ & $.024_{\pm.040}$ & $.053_{\pm.039}$ & $.064_{\pm.068}$ & $.057_{\pm.097}$ & $.066_{\pm.056}$ & $.073_{\pm.071}$ \\
 & \texttt{UPFD}   & $\cellcolor{gray!20}{.157_{\pm.070}}$ & $.041_{\pm.030}$ & $.036_{\pm.034}$ & $.031_{\pm.020}$ & $.036_{\pm.049}$ & $\underline{.062_{\pm.048}}$ & ${.042_{\pm.046}}$ & $.029_{\pm.031}$ \\
  & \texttt{MOTIF}  & $\underline{.382_{\pm.096}}$ & $.270_{\pm.139}$ & $.103_{\pm.090}$ & $.200_{\pm.119}$ & $.313_{\pm.273}$ & $.323_{\pm.185}$ & $\cellcolor{gray!20}{.816_{\pm.169}}$ & ${.323_{\pm.293}}$ \\
 & \texttt{MULTI}  & $\underline{.394_{\pm.086}}$ & $.304_{\pm.157}$ & $.126_{\pm.092}$ & $.195_{\pm.112}$ & $.232_{\pm.217}$ & $.473_{\pm.265}$ & $\cellcolor{gray!20}{.809_{\pm.165}}$ & $.265_{\pm.241}$ \\
\bottomrule
\end{tabular}
}
\label{tab:node_experiment}
\end{table}

\subsubsection{Scalability: Node-level estimation.}
We evaluate our variant, $\text{eXEL}_{\text{node}}$, to verify whether our method can be extended to node-level importance estimation. We use \texttt{IMDB-BINARY} (\texttt{IMDB}), \texttt{ENZYMES}, \texttt{Protein} (\texttt{PROTE}) \cite{tud}, and \texttt{UPFD} \cite{UPFD} for real-world graphs, and \texttt{BAMultiShapes} (\texttt{MULTI}) and \texttt{BA2-Motif} (\texttt{MOTIF}) for synthetic ones. For datasets with available node features, we use the original features; for those without, we adopt one-hot encoding of node degrees \cite{how_powerful}.

As in the graph-level experiments, sparsity is similarly controlled by retaining only the top 30\% of nodes based on the importance scores from each method. 
Table \ref{tab:node_experiment} presents the node-level experiment results, where the scores are averaged across different readout functions. For the fidelity metric, our method ranks first in 12 out of 24 settings (50\%, 4 GNN models $\times$ 6 datasets), and also achieves either first or second place in 22 scenarios (91.7\%). This result demonstrates that \textbf{our variant, $\text{eXEL}_{\text{node}}$, can not only be extended to node-level importance estimation but also consistently achieves notable performance compared to other baselines}.
On synthetic datasets (\texttt{MOTIF} and \texttt{MULTI}), SubgraphX generally performs better than other methods, whereas our $\text{eXEL}_{\text{node}}$ demonstrates strong performance on real-world datasets.

\section{Conclusion and Limitations}

In this paper, we introduced eXEL, a subgraph importance estimation method for pretrained GNNs, formulated as a linear Group Lasso regression problem in the embedding space. Our method is applicable regardless of the output layer type, the readout function, or the ground-truth target label. Through extensive experiments, we demonstrate that our method consistently outperforms existing approaches while effectively leveraging prior domain knowledge of graph substructures through the Group Lasso penalty.

However, our method also has the following limitations. 
First, it reconstructs the graph-level embedding vector, implying that it is applicable only to GNN models equipped with a readout function.
Furthermore, since we adopt the Group Lasso penalty, our method can identify only partitioned subgraphs.
As a future direction, it can be extended to incorporate tree-structured hierarchical subgraph domain knowledge by employing the tree-guided Group Lasso \cite{kim2012tree}.
On the other hand, although the simple Lasso penalty does not require domain prior knowledge, applying it to the final GNN layer may not be appropriate, as it treats nodes independently even though the embeddings have already aggregated information from neighboring nodes through the message-passing mechanism.
Therefore, we plan to enhance our method to a hybrid approach that applies the simple Lasso penalty to earlier GNN layers and the Group Lasso penalty to the later layers.


\begin{credits}
\subsubsection{\ackname} 
This study was supported by the National Research Foundation of Korea (NRF) grant funded by the Ministry of Science and ICT (MSIT) (No. RS-2022-NR068754, No. RS-2025-23525436).
The authors acknowledge the Urban Big data and AI Institute of the University of Seoul supercomputing resources (\url{http://ubai.uos.ac.kr}) made available for conducting the research reported in this paper.
\end{credits}

%
%
\bibliographystyle{splncs04}
\bibliography{reference}

@article{pgm_explainer,
  title={Pgm-explainer: Probabilistic graphical model explanations for graph neural networks},
  author={Vu, Minh and Thai, My T},
  journal={Advances in neural information processing systems},
  volume={33},
  pages={12225--12235},
  year={2020}
}

@article{explain_technique,
  title={Explainability techniques for graph convolutional networks},
  author={Baldassarre, Federico and Azizpour, Hossein},
  journal={arXiv preprint arXiv:1905.13686},
  year={2019}
}

@article{explain_method,
  title={Explainability Methods for Graph Convolutional Neural Networks},
  author={Phillip E. Pope and Soheil Kolouri and Mohammad Rostami and Charles E. Martin and Heiko Hoffmann},
  journal={2019 IEEE/CVF Conference on Computer Vision and Pattern Recognition (CVPR)},
  year={2019},
  pages={10764-10773},

}

@article{gnn_explainer,
  title={GNNExplainer: Generating Explanations for Graph Neural Networks},
  author={Rex Ying and Dylan Bourgeois and Jiaxuan You and Marinka Zitnik and Jure Leskovec},
  journal={Advances in neural information processing systems},
  year={2019},
  volume={32},
  pages={
          9240-9251
        },

}

@inproceedings{subgraphx,
  title={On Explainability of Graph Neural Networks via Subgraph Explorations},
  author={Hao Yuan and Haiyang Yu and Jie Wang and Kang Li and Shuiwang Ji},
  booktitle={International Conference on Machine Learning},
  year={2021},

}

@article{GNN_LRP,
  title={Higher-Order Explanations of Graph Neural Networks via Relevant Walks},
  author={Thomas Schnake and Oliver Eberle and Jonas Lederer and Shinichi Nakajima and Kristof T. Schutt and Klaus-Robert Muller and Gr{\'e}goire Montavon},
  journal={IEEE Transactions on Pattern Analysis and Machine Intelligence},
  year={2020},
  volume={44},
  pages={7581-7596},

}

@article{HGexplainer,
  title={HGExplainer: Explainable Heterogeneous Graph Neural Network},
  author={Grzegorz P. Mika and Amel Bouzeghoub and Katarzyna Wegrzyn-Wolska and Yessin M. Neggaz},
  journal={IEEE International Conference on Web Intelligence and Intelligent Agent Technology},
  year={2023},
  pages={221-229},

}

@article{how_powerful,
  title={How Powerful are Graph Neural Networks?},
  author={Keyulu Xu and Weihua Hu and Jure Leskovec and Stefanie Jegelka},
  journal={ArXiv},
  year={2018},


}

@article{stgraphlime,
  title={Structural Explanations for Graph Neural Networks using HSIC},
  author={Ayato Toyokuni and Makoto Yamada},
  journal={ArXiv},
  year={2023},


}

@article{graph_lime,
  title={GraphLIME: Local Interpretable Model Explanations for Graph Neural Networks},
  author={Q. Huang and Makoto Yamada and Yuan Tian and Dinesh Singh and Dawei Yin and Yi Chang},
  journal={IEEE Transactions on Knowledge and Data Engineering},
  year={2020},
  volume={35},
  pages={6968-6972},

}

@article{relex,
  title={RelEx: A Model-Agnostic Relational Model Explainer},
  author={Yue Zhang and David DeFazio and Arti Ramesh},
  journal={Proceedings of the 2021 AAAI/ACM Conference on AI, Ethics, and Society},
  year={2020},

}

@inproceedings{ig_gnn,
  author       = {Michael Sejr Schlichtkrull and
                  Nicola De Cao and
                  Ivan Titov},
  title        = {Interpreting Graph Neural Networks for NLP With Differentiable Edge
                  Masking},
  booktitle    = {ICLR 2021,
                  },

  year         = {2021},


}

@article{graphxai,
  title={Evaluating explainability for graph neural networks},
  author={Chirag Agarwal and Owen Queen and Himabindu Lakkaraju and Marinka Zitnik},
  journal={Scientific Data},
  year={2022},
  volume={10},

}

@inproceedings{motif_ssl,
  title={Motif-based Graph Self-Supervised Learning for Molecular Property Prediction},
  author={Zaixin Zhang and Qi Liu and Hao Wang and Chengqiang Lu and Chee-Kong Lee},
  booktitle={NeurIPS},
  year={2021},

}

@article{graph_class_survey,
  title={A Comprehensive Survey on Graph Neural Networks},
  author={Zonghan Wu and Shirui Pan and Fengwen Chen and Guodong Long and Chengqi Zhang and Philip S. Yu},
  journal={IEEE Transactions on Neural Networks and Learning Systems},
  year={2019},
  volume={32},
  pages={4-24},

}

@article{chemical_motif,
author = {Degen, Jörg and Wegscheid-Gerlach, Christof and Zaliani, Andrea and Rarey, Matthias},
title = {On the Art of Compiling and Using 'Drug-Like' Chemical Fragment Spaces},
journal = {ChemMedChem},
volume = {3},
number = {10},
pages = {1503-1507},

year = {2008}
}

@inproceedings{pg_expaliner,
author = {Luo, Dongsheng and Cheng, Wei and Xu, Dongkuan and Yu, Wenchao and Zong, Bo and Chen, Haifeng and Zhang, Xiang},
title = {Parameterized explainer for graph neural network},
year = {2020},
booktitle = {NeurIPS},
articleno = {1646},
numpages = {12},
series = {NIPS '20}
}

@article{gin,
  title={How Powerful are Graph Neural Networks?},
  author={Keyulu Xu and Weihua Hu and Jure Leskovec and Stefanie Jegelka},
  journal={ArXiv},
  year={2018},
  volume={abs/1810.00826},

}

@article{lip_bio,
  title={MOGONET integrates multi-omics data using graph convolutional networks allowing patient classification and biomarker identification},
  author={Tongxin Wang and Wei Shao and Zhi Huang and Haixu Tang and J. Zhang and Zhengming Ding and Kun Huang},
  journal={Nature Communications},
  year={2021},
  volume={12},

}

@misc{fidelity,
      title={GraphFramEx: Towards Systematic Evaluation of Explainability Methods for Graph Neural Networks}, 
      author={Kenza Amara and Rex Ying and Zitao Zhang and Zhihao Han and Yinan Shan and Ulrik Brandes and Sebastian Schemm and Ce Zhang},
      year={2024},
      eprint={2206.09677},
      archivePrefix={arXiv},

}

@inproceedings{lip_network,
  title={Blockchain phishing scam detection via multi-channel graph classification},
  author={Zhang, Dunjie and Chen, Jinyin and Lu, Xiaosong},
  booktitle={International conference on blockchain and trustworthy systems},
  pages={241--256},
  year={2021},
  organization={Springer}
}

@inproceedings{lip_text_1,
  title={Graph convolutional networks for text classification},
  author={Yao, Liang and Mao, Chengsheng and Luo, Yuan},
  booktitle={Proceedings of the AAAI conference on artificial intelligence},
  volume={33},
  number={01},
  pages={7370--7377},
  year={2019}
}

@article{lip_bio_1,

  title={Self-supervised graph transformer on large-scale molecular data},
  author={Rong, Yu and Bian, Yatao and Xu, Tingyang and Xie, Weiyang and Wei, Ying and Huang, Wenbing and Huang, Junzhou},
  journal={Advances in neural information processing systems},
  volume={33},
  pages={12559--12571},
  year={2020}
}

@article{lip_hinge,
  title={How do loss functions impact the performance of graph neural networks?},
  author={Gabriel Jonas Duarte and Tamara A. Pereira and Erik Jhones F. do Nascimento and Diego Parente Paiva Mesquita and Amauri Holanda Souza Junior},
  journal={Anais do 15. Congresso Brasileiro de Intelig{\^e}ncia Computacional},
  year={2021},

}

@article{group_lasso,
  title={Model selection and estimation in regression with grouped variables},
  author={Ming Yuan and Yi Lin},
  journal={Journal of the Royal Statistical Society: Series B (Statistical Methodology)},
  year={2006},
  volume={68},

}

@inproceedings{shapley,
  title={Contributions to the theory of games},
  author={Harold W. Kuhn and A. W. Tucker and Melvin Dresher and Philip Wolfe and R. Duncan Luce and H. Frederic Bohnenblust},
  year={1953},

}

@article{hsic,
  title={High-Dimensional Feature Selection by Feature-Wise Kernelized Lasso},
  author={Makoto Yamada and Wittawat Jitkrittum and Leonid Sigal and Eric P. Xing and Masashi Sugiyama},
  journal={Neural Computation},
  year={2012},
  volume={26},
  pages={185-207},

}

@inproceedings{gt_dataset,
  title={Evaluating Attribution for Graph Neural Networks},
  author={Benjam{\'i}n S{\'a}nchez-Lengeling and Jennifer N. Wei and Brian K. Lee and Emily Reif and Peter Wang and Wesley Wei Qian and Kevin McCloskey and Lucy J. Colwell and Alexander B. Wiltschko},
  booktitle={Neural Information Processing Systems},
  year={2020},

}

@article{ogb,
  title={Open graph benchmark: Datasets for machine learning on graphs},
  author={Hu, Weihua and Fey, Matthias and Zitnik, Marinka and Dong, Yuxiao and Ren, Hongyu and Liu, Bowen and Catasta, Michele and Leskovec, Jure},
  journal={Advances in neural information processing systems},
  volume={33},
  pages={22118--22133},
  year={2020}
}

@inproceedings{tdc,
  title={Therapeutics Data Commons: Machine Learning Datasets and Tasks for Drug Discovery and Development},
  author={Kexin Huang and Tianfan Fu and Wenhao Gao and Yue Zhao and Yusuf H. Roohani and Jure Leskovec and Connor W. Coley and Cao Xiao and Jimeng Sun and Marinka Zitnik},
  booktitle={NeurIPS Datasets and Benchmarks},
  year={2021},

}

@article{tud,
  title={TUDataset: A collection of benchmark datasets for learning with graphs},
  author={Christopher Morris and Nils M. Kriege and Franka Bause and Kristian Kersting and Petra Mutzel and Marion Neumann},
  journal={ArXiv},
  year={2020},
  volume={abs/2007.08663},

}

@article{UPFD,
  title={User Preference-aware Fake News Detection},
  author={Yingtong Dou and Kai Shu and Congyin Xia and Philip S. Yu and Lichao Sun},
  journal={Proceedings of the 44th International ACM SIGIR Conference on Research and Development in Information Retrieval},
  year={2021},

}

@article{adaptive_readout,
  title={Graph neural networks with adaptive readouts},
  author={Buterez, David and Janet, Jon Paul and Kiddle, Steven J and Oglic, Dino and Li{\`o}, Pietro},
  journal={Advances in Neural Information Processing Systems},
  volume={35},
  pages={19746--19758},
  year={2022}
}

@article{kim2012tree,
author = {Seyoung Kim and Eric P. Xing},
title = {{Tree-guided group lasso for multi-response regression with structured sparsity, with an application to eQTL mapping}},
volume = {6},
journal = {The Annals of Applied Statistics},
number = {3},
publisher = {Institute of Mathematical Statistics},
pages = {1095 -- 1117},
year = {2012},

}
\end{document}